\documentclass{article}

\PassOptionsToPackage{numbers, compress, sort}{natbib}

\usepackage[final]{neurips_2023}

\usepackage[utf8]{inputenc} 
\usepackage[T1]{fontenc}    
\usepackage{hyperref}       
\usepackage{url}            
\usepackage{booktabs}       
\usepackage{amsfonts}       
\usepackage{nicefrac}       
\usepackage{microtype}      
\usepackage{xcolor}         
\usepackage{makecell}

\usepackage{amsmath,amsfonts,bm}

\def\eqref#1{equation~\ref{#1}}

\def\1{\bm{1}}

\DeclareMathAlphabet{\mathsfit}{\encodingdefault}{\sfdefault}{m}{sl}
\SetMathAlphabet{\mathsfit}{bold}{\encodingdefault}{\sfdefault}{bx}{n}

\usepackage{hyperref}
\usepackage{url}

\usepackage{booktabs} 

\usepackage{graphicx}
\usepackage{float}
\usepackage{amsfonts,amsmath,amssymb,amsthm}
\usepackage{multicol, multirow}
\usepackage{makecell}
\usepackage{adjustbox}
\usepackage{enumitem}
\usepackage{wrapfig}
\usepackage{algorithm}
\usepackage{algorithmic}
\usepackage{afterpage}

\usepackage{cleveref}

\newcommand{\model}{MoleculeJAE}

\usepackage{xcolor}

\definecolor{ForestGreen}{RGB}{34,139,34}

\newcommand{\norm}[1]{\left\lVert#1\right\rVert}
\newcommand{\ie}{\em{i.e.}}
\newcommand{\eg}{\em{e.g.}}

\title{
Molecule Joint Auto-Encoding:\\Trajectory Pretraining with 2D and 3D Diffusion
}

\author{
  Weitao Du\textsuperscript{1,2} \thanks{\texttt{duweitao@mass.ac.cn}} \\
  \And
  Jiujiu Chen\textsuperscript{1,3} \\
  \And
  Xuecang Zhang\textsuperscript{1} \\
  \And
  Zhiming Ma\textsuperscript{2} \\
  \And
  Shengchao Liu\textsuperscript{4} \thanks{\texttt{shengchao.liu@umontreal.ca}}\\
  \And
  \\
  $^1$ Huawei Technologies Co., Ltd. \\
  $^2$ Department of Mathematics, Chinese Academy of Sciences \\
  $^3$ Shenzhen Institute for Advanced Study, University of Electronic Science and Technology of China \\
  $^4$ Department of Computer Science and Operations Research, Université de Montréal \\
}

\begin{document}

\maketitle

\begin{abstract}
Recently, artificial intelligence for drug discovery has raised increasing interest in both machine learning and chemistry domains. The fundamental building block for drug discovery is molecule geometry and thus, the molecule's geometrical representation is the main bottleneck to better utilize machine learning techniques for drug discovery. In this work, we propose a pretraining method for molecule joint auto-encoding (MoleculeJAE). MoleculeJAE can learn both the 2D bond (topology) and 3D conformation (geometry) information, and a diffusion process model is applied to mimic the augmented trajectories of such two modalities, based on which, MoleculeJAE will learn the inherent chemical structure in a self-supervised manner. Thus, the pretrained geometrical representation in MoleculeJAE is expected to benefit downstream geometry-related tasks. Empirically, MoleculeJAE proves its effectiveness by reaching state-of-the-art performance on 15 out of 20 tasks by comparing it with 12 competitive baselines. 
\end{abstract}

\section{Introduction}
\vspace{-1ex}

The remarkable progress in self-supervised learning has revolutionized the fields of molecule property prediction and molecule generation through the learning of expressive representations from large-scale unlabeled datasets~\cite{hu2019strategies,liu2019n,hu2020gpt,sun2019infograph,liu2022molecular,zaidi2022pre,liu2021pre,liu2023group,liu2023symmetry}. Unsupervised representation learning can generally be categorized into two types~\cite{liu2021graph,xie2021self,wu2021self,liu2021self,liu2021pre}: generative-based methods, encouraging the model to encode information for recovering the data distribution; and contrastive-based methods, encouraging the model to learn invariant features from multiple views of the same data.  However, despite the success of large language models like GPT-3 \cite{madotto2020language, zhao2023survey} trained using an autoregressive generation approach , learning robust representations from molecular data remains a significant challenge due to their complex graph structures. Compared to natural language and computer vision data \cite{vahdat2021score,ramesh2022hierarchical}, molecular representations exhibit more complex graph structures and symmetries~\cite{thomas2018tensor}. As molecular dynamics follows the principle of non-equilibrium statistical mechanics~\cite{wood1976molecular}, diffusion models, as a generative method inspired by non-equilibrium statistical mechanics~\cite{sohl2015deep, dockhorn2021score}, are a natural fit for 3D conformation generation. Previous researches have demonstrated the effectiveness of diffusion models, specifically 2D molecular graphs~\cite{jo2022score,vignac2022digress} or 3D molecular conformers~\cite{du2022se,liugroup,shi2021learning,hoogeboom2022equivariant}, for molecular structure generation tasks. However, a crucial question remains: \textit{Can diffusion models be effectively utilized for jointly learning 2D and 3D latent molecular representations?}

\begin{figure}[!t]
    \centering
    \includegraphics[width=1.0\columnwidth]{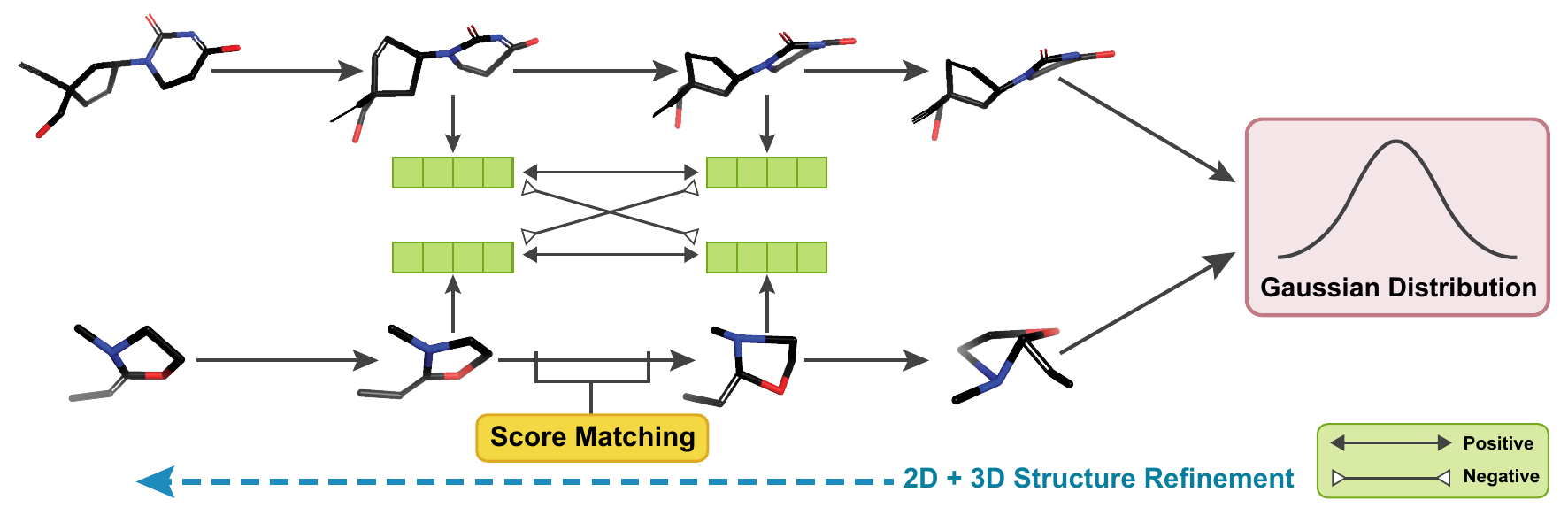}
    \vspace{-3ex}
    \caption{\small Pipeline of \model{}. For each individual molecule, \model{}  utilizes the reconstructive task to perform denoising. For pairwise molecules, it conducts a contrastive learning paradigm to fit the trajectory.}
    \label{fig:pipeline}
    \vspace{-3ex}
\end{figure}


Answering this question is challenging, given that diffusion models are the first generative models based on trajectory fitting. Specifically, diffusion models generate a family of forward random trajectories by either solving stochastic differential equations \cite{song2020score, du2022flexible} or discrete Markov chains \cite{austin2021structured, li2022diffusion, vignac2022digress}, and the generative model is obtained by learning to reverse the random trajectories \cite{ho2020denoising, bansal2022cold}. It remains unclear how the informative representation manifests itself in diffusion models, as compared to other generative models that explicitly contain a semantically meaningful latent representation, such as GAN~\cite{goodfellow2020generative} and VAE~\cite{kingma2013auto}. In addition to the generative power of diffusion models, we aim to demonstrate the deep relationship between the forward process of diffusion models and data augmentation \cite{tian2020makes, sinha2021negative}, a crucial factor in contrastive learning. As a result, we ask whether a trajectory-based self-supervised learning paradigm that leverages both the generative and contrastive benefits of diffusion models can be used to obtain a powerful molecular representation.

 \textbf{Our approach.}
 This work presents \model{} (Molecule Joint Auto-Encoding), a novel trajectory learning framework for molecular representation that captures both 2D (chemical bond structure) and 3D (conformer structure) information of molecules. Our proposed method is designed to respect the $SE(3)$ symmetry of molecule data and is trained by fitting the joint distribution of the data's augmented trajectories extracted from the forward process of the diffusion model. By training the representation in this manner, our framework not only captures the information of real data distribution but also accounts for the corelation between the real data and its noised counterparts. Under certain approximations, this trajectory distribution modeling decomposes into a marginal distribution estimation and a trajectory contrastive regularization task. 
  This multi-task approach yields an effective and flexible framework that can simultaneously handle various types of molecular data, ranging from SE(3)-equivariant 3D conformers to discrete 2D bond connections.
Furthermore, in contrast to diffusion models used for generation tasks that only accept noise as input, most downstream tasks have access to ground-truth molecular structures. To leverage this advantage better, we incorporate an equivariant graph neural network (GNN) block into the architecture, inspired by \cite{nichol2021improved, preechakul2022diffusion}, to efficiently encode crucial information from the ground-truth data. Analogous to conditional diffusion models \cite{nichol2021improved, zhang2023adding}, the encoder's output serves as a meaningful guidence to help the diffused data accurately fitting the trajectory. In summary, our self-supervised learning framework unifies both contrastive and generative learning approaches from a trajectory perspective, providing a versatile and powerful molecular representation that can be applied to various downstream applications.\looseness=-1

Regarding experiments, we evaluate \model{} on 20 well-established tasks drawn from the geometric pretraining literature \cite{zaidi2022pre, liu2021pre}, including the energy prediction at stable conformation and force prediction along the molecule dynamics. Our empirical results support that \model{} can outperform 12 competitive baselines on 15 tasks. We further conduct ablation studies to verify the effectiveness of key modules in \model{}.

\section{Background}
\vspace{-1ex}
In this section, we introduce the diffusion mechanism and relevant notations as a powerful data augmentation framework and elaborate on its instantiation in the context of molecular graph data. 
\subsection{Diffusion Mechanism as a trajectory augmentation}
\vspace{-0.5ex}
The concept of diffusion is widely used in various fields such as physics, mathematics, and computer science. In this paper, we take a broad view of diffusion by defining it as a Markov process with a discrete or continuous time index $t$, starting from a given data point $x_0$ and producing a series of random transformations that lead to $x_t$. If $t \in \{0,1,2,\dots\}$, then $x_t$ is a discrete Markov chain where the probability distribution $p(x_t)$ only depends on the distribution of $x_{t-1}$. We further refer to the conditional probability $p(x_t| x_{t-1})$ as the transition probability of the Markov chain. 
This definition enables us to model the evolution of data over time, allowing us to generate a sequence of augmented data points that span different temporal regimes.

In fact, we can adapt classical data augmentations, such as spatial rotation, color distortion, and Gaussian blurring \cite{yang2022image, trabucco2023effective}, as sources for defining $p(x_t| x_{t-1})$. For instance, Gaussian blurring generates a specific transition probability by setting $p(x_t| x_{t-1}) = \mathcal{N}(x_{t-1}, \epsilon)$, where the scaling factor $\epsilon$ may depend on time $t$.This transition probability rule perturbatively transforms the data for each step, and connecting all steps of transformed data produces a trajectory. 
 In a similar vein, random masking can be seen as a discrete Markov chain, while continuous rotation can be viewed as a deterministic Markov process. By employing this approach, we can effectively expand the scope of one-step data augmentation operations to encompass trajectory-based scenarios. 

In this paper, we build upon the concepts of generalized diffusion processes proposed in \cite{bansal2022cold} to establish a unified framework for the trajectory augmentation examples mentioned earlier. Following the general approach in \cite{bansal2022cold}, each (cold) diffusion model is associated with a degradation Markov process $D(\cdot,t)$ indexed by a time variable $t \in [0,T]$, which introduces perturbations to the initial data $x_0$:\looseness=-1
\begin{equation} \small
\label{noising}
x_0 \xrightarrow{D(\cdot,t)} x_t.\end{equation}

Given that $x_0 \sim p_{data}$, the transformed variable $x_t: = D(x_0,t)$, is also a random variable. To represent the marginal distribution of $D(\cdot,t)$ at time $t$, we use the notation $p_t(\cdot)$. When the context is clear, we will use the shorthand notation $D_t$ to refer to $D(\cdot,t)$. According to Equation (\ref{noising}), we define the pair $(x_0, x_t)$ as a multi-view of the sample $x_0$ at a specific time $t$. It is important to note that, under certain conditions, a Markov process converges to a stationary distribution that effectively erases any memory of the initial data $x_0$. Nevertheless, for small values of $t$, there is a high probability that $x_t$ remains within a ball centered around $x_0$.

In the context of the standard diffusion-based generative model \cite{ho2020denoising, song2020score}, the presence of a reverse process $R_t$ is necessary to undo the effects of the degradation $D_t$. In \cite{song2020score}, the degradation $D(\cdot,t)$ is identified as the solution of a stochastic differential equation (SDE), and it has been demonstrated that a family of SDEs, including the probability flow ordinary differential equation (ODE) proposed in \cite{song2020score,zhang2022gddim}, can reverse the degradation process $D(\cdot,t)$. Specifically, the inversion property implies that $R_{T-t}$ shares the same marginal distribution as $p_t$: $R_{T-t} \sim p_t$. Therefore, as $t$ varies from 0 to $T$, the reverse process $R_t$ gradually restores the data distribution. On the other hand, for cold diffusions, although the model is also trained by reconstructing the data from $D_t$, it does not explicitly assume the existence of a reverse process as rigorously as in the continuous diffusion-based generative model.

\paragraph{Example of heat diffusion}
We use the term 'heat' to describe a diffusion process that involves injecting Gaussian noise. Under this definition, the continuous limit of heat diffusion can be expressed by the solution of stochastic differential equations (SDEs):

\begin{equation}
\label{heat diffusion} \small
dx_t = \mu(x_t, t)dt + \sigma(t)dw_t,\quad t\in [0,T]
\end{equation}

Here, $w_t$ represents the standard Brownian motion. It is worth noting that while the solution $x_t$ of a general SDE may not follow the Gaussian distribution, for the commonly used SDEs such as the Variance Exploding SDE (VE) and Variance Preserving SDE (VP), the solution can be explicitly expressed as:
\begin{equation} \small \label{eq: closed form}
x_t = \alpha(t) x_0 + \beta(t)z,
\end{equation}
where $z$ is sampled from $\mathcal{N}(0,I)$, and both $\alpha(t)$ and $\beta(t)$ are positive scalar functions. Using the forward SDE (\ref{heat diffusion}), we approach the task of data synthesis by gradually denoising the noisy observations $x_t$ to recover $x_0$, which is also accomplished through solving a reverse SDE. For the general formula and further details, readers can refer to \cite{ho2020denoising, song2020score}. One fact we will use later is that the reverse SDE fundamentally relies on the score function $\nabla_x \log p_t (x)$, which serves as the parameterization objective in the score matching framework \cite{song2020score, huang2021variational}.

\paragraph{Example of discrete (cold) diffusion}
In contrast to the continuous heat diffusion framework, we now introduce a family of discrete Markov Chains as an instance of discrete (cold) diffusion. This example is particularly suitable for modeling categorical data. In the unconditional setting, the transition probability of the discrete diffusion at time step $t$ is defined as:
\begin{equation}
\label{absorbing} \small
q(x_t | x_{t-1}) = \text{Multinomial}(x_t, p = x_{t-1} Q_t),
\end{equation}
where $Q_t$ represents a Markov transition matrix. An illustrative example is the \textbf{absorbing diffusion}, as introduced by \cite{austin2021structured}, where
$$Q_t = (1 - \beta_t) I + \beta_t e_m^T,$$
and $e_m$ is a one-hot vector with a value of 1 at the absorbing state $m$ and zeros elsewhere. Typically, the state $m$ is chosen as the 'masked' token. Hence, absorbing diffusion can be regarded as a masking process with a dynamical masking ratio $\beta_t$. Similarly to the heat diffusion, the corresponding reverse process can also be formulated as a discrete diffusion. However, the training objective shifts from fitting the score function (which only exists in the continuous case) to directly fitting the 'reconstruction probability' $p(x_0 | x_t)$.

\subsection{Graph-based molecular representation} \label{sec: Graph-based molecular representation}
\vspace{-0.5ex}
In this paper, our primary focus is on the graph representation of molecules. Let $G = (V,E,P, H)$ denote the integrated molecular graph, consisting of $n:=|V|$ atoms and $|E|$ bonds. 
The matrix $P \in \mathbf{R}^{n\times 3}$ is utilized to represent the 3D conformer of the molecule, containing the positions of each node. Moreover, within the graph $G$, the edge collection $E$ represents the graphical connections and chemical bond types between atoms. Additionally, we have the node-wise feature matrix $H \in \mathbf{R}^{n\times h}$, where, for the purposes of this article, we consider the formal charges and atom types as the components of $H$. In the method section, we will demonstrate how to build trajectories based on $G$.\looseness=-1

\section{Method} \label{sec:method}
\vspace{-1ex}
Now we illustrate the process of obtaining molecule augmented trajectories, based on which, \model{} is proposed to estimate the trajectory distribution. In Section \ref{sec:trajectory_construction}, we will outline the construction of equivariant molecular trajectories. Subsequently, in Section \ref{sec:trajectory_learning}, we will introduce our theoretical self-supervised learning framework for these trajectories. Our hypothesis is that a good representation should encode the information from the distribution of augmented trajectories. This hypothesis forms the practical basis of our core reconstructive and contrastive loss, which will be discussed in Section \ref{sec:reconstructive_and_contrastive_loss}.  Finally, in Section \ref{sec:model_architecture}, we will present the key architectures. A comprehensive discussion of the related works is in Appendix B.

\subsection{Equivariant Molecule Trajectory Construction}\label{sec:trajectory_construction}
\vspace{-0.5ex}
In the field of molecular representation learning, our objective is to jointly estimate the distribution of a molecule's 2D topology (including atom types and chemical bonds) and its 3D geometries (conformers). Building upon the notations in Section \ref{sec: Graph-based molecular representation}, our goal is to construct \textbf{augmented} trajectories $x_t := (H(t), E(t), P(t))$ from $G$. The challenging aspect lies in preserving the $SE(3)$ symmetry of the position matrix $P$, which we will describe in detail below.

Given two 3D point clouds (molecular conformers) $P_1$ and $P_2$, we say they are SE(3)-isometric if there exists an $R \in SE(3)$ such that $P_1 = RP_2$. In probabilistic terms, let $p_{3D}(x_{3D})$ be the probability density of all 3D conformers denoted by $\mathcal{C}$. Isometric conformers necessarily have the same density, i.e.,
\begin{equation}     
\label{eq: 3d density}
p_{3D}(x_{3D}) = p_{3D}(\textbf{R}x_{3D}),\ \ \ \forall \  \textbf{R} \in SE(3), x_{3D} \in \mathcal{C}.    
\end{equation}
 Utilizing the symmetry inductive bias has been shown \cite{satorras2021n,du2022se} to greatly reduce the complexity of data augmentation. Additionally, in order to respect the $SE(3)$ symmetry, the augmented trajectory $P(0) \rightarrow P(t)$ should also be SE(3)-equivariant, satisfying
$$\textbf{R} P(x_{3D}) = P(\textbf{R}x_{3D}),\   \ \ \forall \  \textbf{R} \in SE(3).$$
This condition imposes a rigorous restriction on the form of the forward SDE (\ref{heat diffusion}) when it is applied to generate the augmented trajectory of conformers. However, if we constrain the form of $x_t$ to Eq. \ref{eq: closed form}, the $SE(3)$ equivariance is automatically satisfied due to the $SE(3)$ equivariance of both the Gaussian random variable $z$ and the original data $x_0$. We leave the formal proof in Appendix A. 

Regarding the 2D component, let $x_{2D}(t) := (H(t), E(t))$, where $x_{2D}$ consists of invariant scalars that remain unchanged under $SE(3)$ transformations. Although $x_{2D}$ is categorical in nature, we can treat them as continuous scalars that can be perturbed by Gaussian noise. During inference (generation), these continuous scalars are quantized back into categorical integers (see \cite{jo2022score} for details). This allows both $x_{2D}(t)$ and $P(t)$ to undergo heat diffusion. By combining the 2D and 3D parts, we introduce the system of SDEs for $x_t$:
\begin{equation} \label{joint sde}
\small
\begin{cases}
d P(t) = -P(t)dt + \sigma_1(t)dw_t^1, &  \\
dH(t) = \mu_2(H(t), t)dt + \sigma_2(t)dw_t^2, &\\  
dE(t) = \mu_3(E(t), t)dt + \sigma_3(t)dw_t^3. & 
\end{cases}
\end{equation}
It is worth mentioning that although the components of Eq. \ref{joint sde} are disentangled, the corresponding reverse diffusion process and its score function are entangled. A common choice of $\mu(x,t)$ is $\mu(x,t): = -x$,
then will utilize the explicit solution of equations (\ref{joint sde}) to generate the molecular augmentation trajectory. Additionally, it is also possible to perform equivariant diffusion of the 2D and 3D joint representation through equivariant cold diffusion, following the approach described in (\ref{absorbing}). The detailed framework for this approach is provided in the Appendix A.

\subsection{Equivariant Molecule Trajectory Learning with Density Auto-Encoding} \label{sec:trajectory_learning}
To provide a formal foundation for self-supervised learning from augmented trajectories, we revisit the concept of denoising from a trajectory perspective. According to the Kolmogorov extension theorem \cite{chmiela2017machine}, every stochastic process defines a probabilistic measure on the trajectory space, uniquely determined by a set of finite-dimensional joint distributions that satisfy consistency conditions. Therefore, estimating the probability of the infinite-dimensional trajectory space is equivalent to estimating the joint probabilities $p(x_{t_1},\dots, x_{t_k})$ for each finite time sequence $t_{1},\dots ,t_{k}\in [0,T]$. From the standpoint of data augmentation, our specific focus is on learning the joint distribution of trajectory augmentation pairs: $p(x_0,x_t)$ (solution of Eq. \ref{joint sde}), which determines how the noisy $x_t$ corresponds to the original (denoised) $x_0$.

To differentiate our notion of "Auto-Encoding" from the "denoising" method utilized in DDPM~\cite{ho2020denoising} (denoising diffusion probabilistic models), it is important to highlight that traditional diffusion-based generative models are based on marginal distribution modeling. In this framework, joint distributions induce marginal distributions, but not vice versa. This distinction becomes more apparent in the continuous SDE formalism of diffusion models \cite{song2020score}, where multiple denoising processes from $x_t$ to $x_0$ can be derived while still sharing the same marginal distributions. However, the joint distributions of a probabilistic ODE flow significantly differ from those of the reverse SDE (as shown in Eq. \ref{eq: family}) due to the deterministic nature of state transitions in an ODE between timesteps. In the following, we formally demonstrate this observation from the perspective of maximizing the trajectories' joint probabilistic log-likelihood.

For a given time $t \in [0,T]$, let us assume that the probability of the random pair $(x_0,x_t)$ defined in Eq. \ref{noising} is determined by a joint density function $p(x_0,x_t)$. Our objective is to approximate $p(x_0,x_t)$ within a variational class $p_{\theta}(x_0,x_t)$. Maximizing the likelihood of this joint distribution leads to the following optimization problem:
\begin{equation} \label{argmax} \small
\text{argmax}_{\theta} \prod_{i=1}^{n} p_{\theta}(x_0^i,x_t^i),\end{equation}
where $\{(x_0^i,x_t^i)\}_{i=1}^n$ represents the collection of $n$ augmented samples from the training dataset.
Following the tradition of Bayesian inference \cite{goan2020bayesian}, we parameterize   $p_{\theta}$ by defining a joint energy function $\textbf{E}_{\theta}(x_0, x_t)$ such that:
$p_{\theta}(x_0,x_t) = \frac{1}{Z_{\theta}}e^{- \textbf{E}_{\theta}(x_0, x_t)},$
where $Z_{\theta}(t)$ is the intractable normalization constant depending on $t$. Therefore, the maximal likelihood framework reduces to solving
$$\text{argmin}_{\theta}\mathbb{E}_{p(x_0, x_t)} \left[\textbf{E}_{\theta}(x_0, x_t)\right].$$
However, directly optimize $p_{\theta}(x_0,x_t)$ by taking the gradient with respect to $\theta$ is challenging because $\frac{\partial \log p_{\theta}(x_0, x_t)}{\partial \theta}$ contains an intractable term:
$\frac{\partial Z_{\theta}(t)}{\partial \theta}$. To circumvent this issue, we borrow ideas from \cite{grathwohl2019your, zhao2020joint} by treating the transformed $x_t$ as an infinitely dimensional "label" of $x_0$. More precisely, we consider the parameterized marginal density $q_{\theta} (x_0)$ as:
\begin{equation} \label{marginal} \small
q_{\theta} (x_0) := \int p_{\theta}(x_0,x_t) dx_t,\end{equation}
which involves integrating out $x_t$. We define the marginalized energy function with respect to $x_0$ as:
$\small \bar{\textbf{E}}_{\theta} (\cdot) := - \log  \int \exp (-\textbf{E}_{\theta}(\cdot, x_t)) d x_t .$
Now, let $f_{\theta}(x_0 , x_t)$ denote the normalized conditional density function $p_{\theta}(x_t | x_0)$, we have
$\small \frac{\partial \log f_{\theta}(x_t | x_0)}{\partial \theta} = - \frac{\partial \textbf{E}_{\theta}(x_0, x_t)}{\partial \theta} + \frac{\partial \bar{\textbf{E}}_{\theta} (x_0)}{\partial \theta}.$
By taking the empirical expectation with respect to the finitely sampled pair $(x_0,x_t) \sim p(x_0,x_t)$, we can decompose the gradient of the maximum likelihood as follows (see Appendix A for the full derivation):
\begin{equation} \label{decomposition} \small
\Tilde{\mathbb{E}}_{p(x_0,x_t)}\left[\frac{\partial \log p_{\theta}(x_0, x_t)}{\partial \theta} \right]=  \Tilde{\mathbb{E}}_{p(x_0)}\left[\frac{\partial \log q_{\theta}(x_0)}{\partial \theta}\right] + \Tilde{\mathbb{E}}_{p(x_0,x_t)}\left[\frac{\partial \log f_{\theta}(x_0 , x_t)}{\partial \theta}\right] ,\end{equation}
here we use $\Tilde{\mathbb{E}}$ to denote the expectation with respect to the empirical expectation. 
Note that this decomposition holds for $(x_s,x_t)$ for any two different time steps ($0 \leq s,t \leq T$) of the trajectories. 
In the next section, we decompose~\Cref{decomposition} into two sub-tasks, leading to our goal of optimizing these two parts simultaneously, as discussed below.

\subsection{Reconstructive and Contrastive Tasks} \label{sec:reconstructive_and_contrastive_loss}
\vspace{-0.5ex}
In what follows, we denote the latent representation of data $x$ by $h_{\theta}(x)$  (the equivariant model for building $h_{\theta}$ will be discussed in the next section). Based on Eq. \ref{decomposition}, we introduce two tasks for training $h_{\theta}(x)$ that stem from this decomposition.

\textbf{Reconstructive task.\ } 
The first term $\Tilde{\mathbb{E}}_{p(x_0)}\left[\frac{\partial \log q_{\theta}(x_0)}{\partial \theta}\right]$ in Eq. \ref{decomposition} aims to reconstruct the distribution of data samples. Therefore,  we refer this term as the \textbf{reconstruction} task, which involves modeling the marginal distribution $p_{data}(x_0)$ using $q_{\theta}(x_0)$.

Although it is possible to directly train the likelihood reconstruction term in the auto-regressive case, such as using the noise conditional maximum likelihood loss $\mathbb{E}_{t \sim [0,T]} \mathbb{E}_{x_t \sim p_t} \log q_{\theta}(x_t)$ proposed in \cite{li2022autoregressive}, we instead adopt trajectory score matching \cite{song2020score}, which is applicable for non-autoregressive graph data. The score matching loss is defined as follows:
\begin{equation} \label{score matching}
\mathcal{L}_{sc}: = \mathbb{E}_{t \sim [0,T]} \mathbb{E}_{p(x_0)p(x_t|x_0)} [\lVert \nabla \log p (x_t | x_0) - s_{\theta} (x_t, t)\rVert^2].
\end{equation}
We choose this approach for two reasons: First, training the score function enables the model to generate new samples by solving the reverse process, thus facilitating generative downstream tasks. Second, when $t=0$, the score function for 3D structures can be interpreted as a "pseudo" force field for the molecular system \cite{zaidi2022pre}, containing essential information about the molecular dynamics.
Furthermore, \cite{huang2021variational} provided a rigorous proof demonstrating that score matching formula \ref{score matching} serves as a variational lower bound for the marginal log-likelihood $\log q_{\theta} (x_0)$. This theoretical guarantee solidifies the effectiveness of score matching as a training objective. 

\begin{figure}[!t]
    \centering
    \includegraphics[width=1.0\columnwidth]{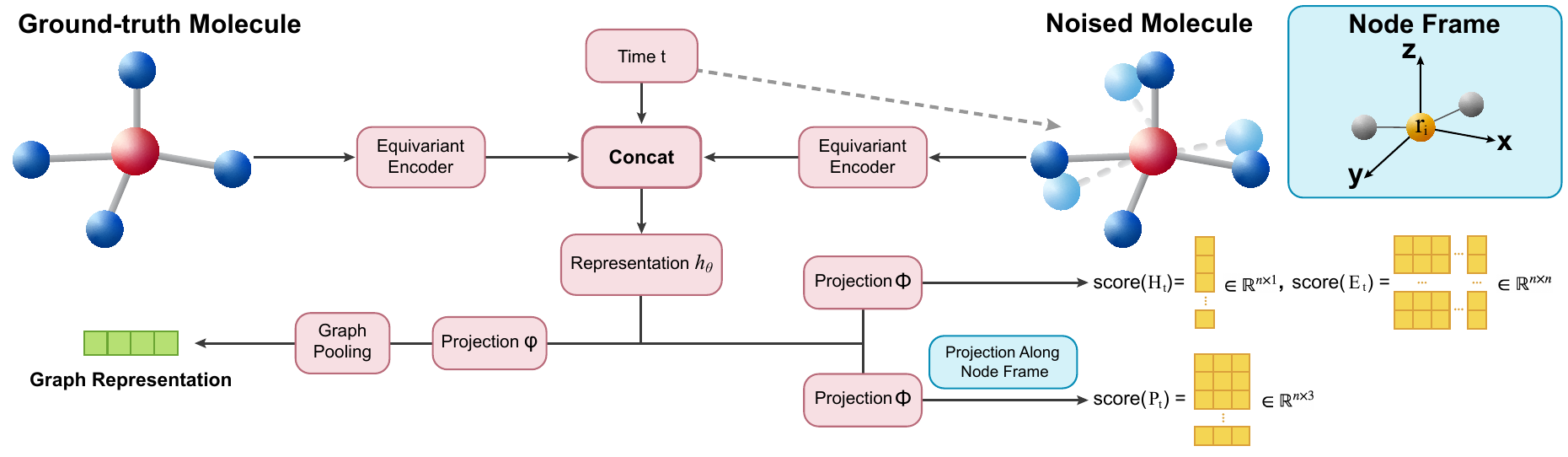}
    \vspace{-3ex}
    \caption{\small Architecture of \model{}. The inputs are ground-truth molecules with both 2D and 3D structures. \model{} also adopts the noised inputs denoising, so as to model the trajectory distribution. The outputs are three score functions for conformer and bond representations, which flow into the general pipeline in~\Cref{fig:pipeline}.}
    \label{fig:art}
    \vspace{-3ex}
\end{figure}

For molecule trajectories, the score function encompasses both the 2D and 3D components. Moreover, the SE(3)-invariant density $p_{3D}$ defined by Eq. \ref{eq: 3d density} implies that the corresponding 3D score function $\nabla_{x} p_{3D}(x)$  (represented as $\text{score}(P_t)$ in Fig. \ref{fig:art}) is equivariant under $SE(3)$:
$$
\small \nabla_{x} p_{3D}(\textbf{R}x) = \textbf{R} \nabla_{x} p_{3D}(x),\ \ \ \forall \  \textbf{R} \in SE(3),\ \  x \in \mathcal{C}.$$
In conclusion, the symmetry principle mandates that the score neural network takes an equivariant vector field (representing the positions of all atoms) and invariant atom features as input. It then produces two score functions that adhere to different transformation rules:
1. $\nabla_{x} p_{3D}(x)$: SE(3)-equivariant;
2. $\nabla_{y} p_{2D}(y)$: SE(3)-invariant.

\textbf{Contrastive task.\ }
We have demonstrated that optimizing the score matching task allows us to capture information about the marginal distributions of the trajectories. However, the joint distribution contains additional valuable information. As an example, consider the following parameterized random processes, all of which share the same marginal distributions but exhibit varying joint distributions (as proven in \cite{huang2021variational}):
\begin{equation} \label{eq: family}
\small
dy_t = [f(y_t,t) - \frac{1+\lambda^2}{2}g^2(t) s_{\theta}(y_t,t)]dt + \lambda g(t)dB_t,\end{equation}
 for $\lambda > 0$. 
Hence, it is theoretically necessary to optimize the second term  $\Tilde{\mathbb{E}}_{p(x_0,x_t)}\left[\frac{\partial \log f_{\theta}(x_0 , x_t)}{\partial \theta}\right]$ of Eq. \ref{decomposition}.
By employing the conditional probability formula, we have:
\begin{equation} \label{conditional}
\small
f_{\theta}(x_0 , x_t) = \frac{p_{\theta}(x_0, x_t)}{\int p_{\theta}(x_0, y) dy}. 
\end{equation}
It is important to note that the troublesome normalizing constant $Z_{\theta} (t)$ is cancelled out in Eq. \ref{conditional}. In practice, the integral  $\int p_{\theta}(x_0, y) dy$ is empirically approximated using Monte Carlo sampling. To make a connection with contrastive learning (CL), recall that in CL, a common approach is to align the augmented views ($(x,x^+)$) of the same data and simultaneously contrast the augmented views of different data ($x^-$).  By treating the joint distribution as a similarity measure, Eq. \ref{conditional} can be seen as implicitly imposing two regularization conditions on $f_{\theta}(x_0,x_t)$:  ensuring a high probability for $p(x_0,x_t)$ and a low probability for $p(x_0,y)$. This notion of similarity motivates us to refer to maximizing the second term of Eq. \ref{decomposition} as a \textbf{contrastive} task.

\paragraph{Contrastive surrogate} 
However, estimating $p_{\theta}(x_0,x_t)$ is challenging due to the intractability of closed-form joint distributions. To overcome this difficulty, we propose a black-box surrogate $\varphi$ that represents the mapping from the latent representation $h_{\theta}(x_t)$ to $p_{\theta}(x_0,x_t)$ (following the notation in figure \ref{fig:art}). Specifically, let $(h_{\theta}(x_0),h_{\theta}(x_t))$ denote the representation pair obtained from the input molecule data $(x_0,x_t)$. Then, the surrogate of $p_{\theta}(x_0,x_t)$ is defined by $p_{\theta}(x_0,x_t) = \frac{1}{Z(t)}\exp \{- \frac{\lVert \varphi(h_{\theta}(x_0)) -  \varphi(h_{\theta}(x_t))\rVert^2}{\tau^2(t)}+C(x_0)\}.$ Here, $\tau(t)$ is a monotone annealing function with respect to $t$. By using Eq. \ref{conditional}, the unknown normalization constant $Z(t)$ and $C(x_0)$ cancel out, resulting in the following approximation:
\begin{equation}  \label{contrastive}  
f_{\theta}(x_0, x_t)  \approx  \frac{\exp \{- \lVert \varphi(h_{\theta}(x_0)) - \varphi(h_{\theta}(x_{t}))\rVert^2\}}{\int \exp \{- \lVert \varphi(h_{\theta}(x_0)) - \varphi(h_{\theta}(y))\rVert^2\} dy }.\end{equation}
Our surrogate is reasonable because our modeling target $p(x_0,x_t)$ is derived from the joint distributions of a (continuous) Markov process. When $x_t$ lies along the augmented trajectory of a specific data sample $x_0$ and approaches $x_0$ as $t \rightarrow 0$, the log-likelihood  $\log p_{\theta}(x_0,x_t)$ is also achieves a local maximum. Therefore, based on the Taylor expansion, the leading non-constant term takes the form of a quadratic expression.

\paragraph{Final Objective}
By combining the reconstruction and contrastive regularization tasks, we define the final \textbf{multi-task} training objective of \model{} as a weighted sum of the two tasks:
\begin{equation} \label{eq:final_objective}
\small
\lambda_1 \mathcal{L}_{sc} + \lambda_2 \mathcal{L}_{co},
\end{equation}
where $\lambda_1, \lambda_2$ are weighting coefficients, and
$\mathcal{L}_{co} : = \mathbb{E}_{t \sim [0,T]} \mathbb{E}_{(x_0,x_t) \sim p(x_0,x_t)} f_{\theta} (x_o,x_t).$
In Fig. \ref{fig:art}, the noised input $x_t$ undergoes an encoding process to obtain the latent representation $h_{\theta}(x_t)$, which is then split into two branches:
\begin{enumerate}
    \item The first branch passes through a score neural network $\phi$ for reconstruction: $s_{\theta}(x_t) = \phi (h_{\theta}(x_t))$;
    \item The second branch incorporates the original data representation  $h_{\theta}(x_0)$ and further projects the pair  $(h_{\theta}(x_0), h_{\theta}(x_t) )$ through a non-linear projection head $\varphi$ for contrastive learning.
\end{enumerate}
Note that the black-box projection function $g$ (although not used in downstream tasks) also participates in the pretraining optimization, following the conventional contrastive learning frameworks \cite{chen2020simple, gupta2022understanding}. See Fig. \ref{fig:pipeline} for a graphical illustration of \model{}'s pipeline. 



\subsection{Model Architecture of \model{}} \label{sec:model_architecture}
\vspace{-0.5ex}
In pursuit of a meaningful latent molecular representation, we design our joint auto-encoder model as a conditional diffusion based model, inspired by \cite{meng2021sdedit, nichol2021improved, preechakul2022diffusion}. In our model, the condition does not come from labels, but rather an encoding of the ground-truth molecule. Specifically, to obtain the latent representation $h_{\theta}$ shown in Fig. \ref{fig:art}, we implement two \textbf{equivariant} encoders that satisfy the $SE(3)$ symmetry proposed in Section \ref{sec:trajectory_construction}. One encoder takes the original molecule as input, and its output is used as a condition to help the other encoder that encodes the noisy molecule for reconstruction. The only requirement for the architecture of the two encoders is that the output should be invariant. Therefore, any $SE(3)$ equivariant GNN \cite{du2023new} that outputs invariant scalars will suffice.

\textbf{Equivariant decoder.\ } With our representation $h_{\theta}$ which depends on $(x_0,x_t,t)$, the decoder part of MoleculeJAE is divided into two heads. One head is paired with node-wise SE(3) frames to match the $SE(3)$ equivariant score function, while the other head generates an $SE(3)$ invariant representation that is used for contrastive learning (see Fig. \ref{fig:art} for a complete illustration). Further details on the model design are left in Appendix A.

\section{Experiment}
\vspace{-1ex}

\begin{table}[tb]
\setlength{\tabcolsep}{5pt}
\fontsize{9}{9}\selectfont
\caption{
\small
Results on 12 quantum mechanics prediction tasks from QM9. We take 110K for training, 10K for validation, and 11K for testing. The evaluation is mean absolute error, and the best and the second best results are marked in \underline{\textbf{bold}} and \textbf{bold}, respectively.
}
\vspace{-1.5ex}
\label{tab:main_QM9_result}
\begin{adjustbox}{max width=\textwidth}
\begin{tabular}{l c c c c c c c c c c c c}
\toprule
Pretraining & $\alpha$ $\downarrow$ & $\nabla \mathcal{E}$ $\downarrow$ & $\mathcal{E}_\text{HOMO}$ $\downarrow$ & $\mathcal{E}_\text{LUMO}$ $\downarrow$ & $\mu$ $\downarrow$ & $C_v$ $\downarrow$ & $G$ $\downarrow$ & $H$ $\downarrow$ & $R^2$ $\downarrow$ & $U$ $\downarrow$ & $U_0$ $\downarrow$ & ZPVE $\downarrow$\\
\midrule

-- (random init) & 0.060 & 44.13 & 27.64 & 22.55 & 0.028 & 0.031 & 14.19 & 14.05 & 0.133 & 13.93 & 13.27 & 1.749\\

Type Prediction & 0.073 & 45.38 & 28.76 & 24.83 & 0.036 & 0.032 & 16.66 & 16.28 & 0.275 & 15.56 & 14.66 & 2.094\\

Distance Prediction & 0.065 & 45.87 & 27.61 & 23.34 & 0.031 & 0.033 & 14.83 & 15.81 & 0.248 & 15.07 & 15.01 & 1.837\\

Angle Prediction & 0.066 & 48.45 & 29.02 & 24.40 & 0.034 & 0.031 & 14.13 & 13.77 & 0.214 & 13.50 & 13.47 & 1.861\\

3D InfoGraph & 0.062 & 45.96 & 29.29 & 24.60 & 0.028 & 0.030 & 13.93 & 13.97 & 0.133 & 13.55 & 13.47 & 1.644\\

GeoSSL-RR & 0.060 & 43.71 & 27.71 & 22.84 & 0.028 & 0.031 & 14.54 & 13.70 & \textbf{0.122} & 13.81 & 13.75 & 1.694\\

GeoSSL-InfoNCE & 0.061 & 44.38 & 27.67 & 22.85 & \textbf{0.027} & 0.030 & 13.38 & 13.36 & \underline{\textbf{0.116}} & 13.05 & 13.00 & 1.643\\

GeoSSL-EBM-NCE & 0.057 & 43.75 & 27.05 & 22.75 & 0.028 & 0.030 & 12.87 & 12.65 & 0.123 & 13.44 & 12.64 & 1.652\\

3D InfoMax & 0.057 & \textbf{42.09} & 25.90 & 21.60 & 0.028 & 0.030 & 13.73 & 13.62 & 0.141 & 13.81 & 13.30 & 1.670\\

GraphMVP & 0.056 & \underline{\textbf{41.99}} & 25.75 & \textbf{21.58} & \textbf{0.027} & \textbf{0.029} & 13.43 & 13.31 & 0.136 & 13.03 & 13.07 & 1.609\\

GeoSSL-DDM-1L & 0.058 & 42.64 & 26.32 & 21.87 & 0.028 & 0.030 & 12.61 & 12.81 & 0.173 & 12.45 & 12.12 & 1.696\\

GeoSSL-DDM & 0.056 & 42.29 & \underline{\textbf{25.61}} & 21.88 & \textbf{0.027} & \textbf{0.029} & \textbf{11.54} & \textbf{11.14} & 0.168 & \textbf{11.06} & \underline{\textbf{10.96}} & 1.660\\
\midrule

\model{} & \underline{\textbf{0.056}} & 42.73 & 25.95 & \underline{\textbf{21.55}} & \underline{\textbf{0.027}} & \underline{\textbf{0.029}} & \underline{\textbf{11.22}} & \underline{\textbf{10.70}} & 0.141 & \underline{\textbf{10.81}} & \underline{\textbf{10.70}} & \underline{\textbf{1.559}}\\

\bottomrule
\end{tabular}
\end{adjustbox}
\vspace{-2ex}
\end{table}

\begin{table}[tb]
\setlength{\tabcolsep}{5pt}
\fontsize{9}{9}\selectfont
\centering
\caption{
\small
Results on eight force prediction tasks from MD17. We take 1K for training, 1K for validation, and 48K to 991K molecules for the test concerning different tasks. The evaluation is mean absolute error, and the best results are marked in \underline{\textbf{bold}} and \textbf{bold}, respectively.
}
\label{tab:main_MD17_result}
\vspace{-1.5ex}
\begin{adjustbox}{max width=\textwidth}
\begin{tabular}{l c c c c c c c c}
\toprule
Pretraining & Aspirin $\downarrow$ & Benzene $\downarrow$ & Ethanol $\downarrow$ & Malonaldehyde $\downarrow$ & Naphthalene $\downarrow$ & Salicylic $\downarrow$ & Toluene $\downarrow$ & Uracil $\downarrow$ \\
\midrule

-- (random init) & 1.203 & 0.380 & 0.386 & 0.794 & 0.587 & 0.826 & 0.568 & 0.773\\


Type Prediction & 1.383 & 0.402 & 0.450 & 0.879 & 0.622 & 1.028 & 0.662 & 0.840\\

Distance Prediction & 1.427 & 0.396 & 0.434 & 0.818 & 0.793 & 0.952 & 0.509 & 1.567\\

Angle Prediction & 1.542 & 0.447 & 0.669 & 1.022 & 0.680 & 1.032 & 0.623 & 0.768\\

3D InfoGraph & 1.610 & 0.415 & 0.560 & 0.900 & 0.788 & 1.278 & 0.768 & 1.110\\

GeoSSL-RR & 1.215 & 0.393 & 0.514 & 1.092 & 0.596 & 0.847 & 0.570 & 0.711\\

GeoSSL-InfoNCE & 1.132 & 0.395 & 0.466 & 0.888 & 0.542 & 0.831 & 0.554 & 0.664\\

GeoSSL-EBM-NCE & 1.251 & 0.373 & 0.457 & 0.829 & 0.512 & 0.990 & 0.560 & 0.742\\

3D InfoMax & 1.142 & 0.388 & 0.469 & 0.731 & 0.785 & 0.798 & 0.516 & 0.640\\

GraphMVP & \textbf{1.126} & 0.377 & 0.430 & 0.726 & \underline{\textbf{0.498}} & 0.740 & 0.508 & 0.620\\

GeoSSL-DDM-1L & 1.364 & 0.391 & 0.432 & 0.830 & 0.599 & 0.817 & 0.628 & 0.607\\

GeoSSL-DDM & \underline{\textbf{1.107}} & \textbf{0.360} & \underline{\textbf{0.357}} & 0.737 & 0.568 & 0.902 & \textbf{0.484} & \textbf{0.502}\\

\midrule

\model{} & 1.289 & \underline{\textbf{0.345}} & \textbf{0.365} & \underline{\textbf{0.613}} & \underline{\textbf{0.498}} & \underline{\textbf{0.712}} & \underline{\textbf{0.480}} & \underline{\textbf{0.463}}\\

\bottomrule
\end{tabular}
\end{adjustbox}
\vspace{-3ex}
\end{table}

\subsection{\model{} Pretraining}
\textbf{Dataset.}
For pretraining, we use PCQM4Mv2~\cite{hu2020ogb}. It extracts 3.4 million molecules from PubChemQC~\cite{nakata2017pubchemqc} with both 2D topology and 3D geometry.

\textbf{Backbone models.}
We want to highlight that \model{} is agnostic to backbone geometric GNNs. In this work, we follow previous works in using SchNet model~\cite{schutt2018schnet} for 3D conformation. For the 2D GNN representation, we take a simple version by mainly modeling the bond information (details in~\Cref{sec:app:model_and_implementation_details}). For the SDE models~\cite{song2020score} for generating the joint trajectories, we consider both VE and VP, as in Eq. \ref{joint sde}.

\textbf{Baselines for 3D conformation pretraining.}
Recently, few works have started to explore 3D conformation pretraining. For instance, GeoSSL~\cite{liu2022molecular} provides comprehensive baselines.
The initial three baselines involve type prediction, distance prediction, and angle prediction, respectively aiming to predict the masked atom type, pairwise distance, and triplet angle.
The next baseline is 3D InfoGraph. It is a contrastive SSL method and predicts whether the node- and graph-level 3D representation are for the same molecule.
Last but not least, GeoSSL proposes a new SSL family on geometry:
GeoSSL-RR, GeoSSL-InfoNCE, and GeoSSL-EBM-NCE are to maximize the MI between the conformation and augmented conformation using different objective functions, respectively. GeoSSL-DDM optimizes the same objective function using denoising distance matching.
GeoSSL-DDM-1L~\cite{zaidi2022pre} is a special case of GeoSSL-DDM with one layer of denoising.

\textbf{Baselines for 2D and 3D multi-modal pretraining.}
Additionally, several works have utilized both 2D topology and 3D geometry modalities for molecule pretraining. Vanilla GraphMVP~\cite{liu2021pre} utilizes both the contrastive and generative SSL, and 3D InfoMax~\cite{stark20223d} only uses the contrastive learning part.

In the following, we provide the downstream tasks for applying our pre-trained MoleculeJAE. The experiment on joint generation of the 2D and 3D structures of molecules are provided in Appendix \ref{appendix: generation}.

\subsection{Quantum Property Prediction}
\vspace{-0.5ex}
QM9~\cite{ramakrishnan2014quantum} is a dataset of 134K molecules, consisting of nine heavy atoms. It contains 12 tasks, which are related to the quantum properties. Among 12 tasks, the tasks related to the energies are very important, {\eg}, $U$ and $U_0$ are the internal energies at 0K and 298.15K, respectively. The other two energies, $H$ and $G$ can be obtained from $U$ accordingly.
The main results are in~\Cref{tab:main_QM9_result}. We can observe that \model{} can outperform 12 baselines on 9 out of 12 tasks. We want to highlight that these baselines are very competitive, and certain works ({\eg}, GeoSSL) also model the 3D trajectory. Noticeably, \model{} can reach the best performance on four energy-related tasks.

\begin{table}[tb]
\setlength{\tabcolsep}{5pt}
\fontsize{9}{9}\selectfont
\caption{
\small
Ablation studies of contrastive loss term in \model{}. The ablation results are on QM9.
}
\vspace{-1.5ex}
\label{tab:ablation_contrastive_loss_QM9_result}
\begin{adjustbox}{max width=\textwidth}
\begin{tabular}{l c c c c c c c c c c c c}
\toprule
Pretraining & $\alpha$ $\downarrow$ & $\nabla \mathcal{E}$ $\downarrow$ & $\mathcal{E}_\text{HOMO}$ $\downarrow$ & $\mathcal{E}_\text{LUMO}$ $\downarrow$ & $\mu$ $\downarrow$ & $C_v$ $\downarrow$ & $G$ $\downarrow$ & $H$ $\downarrow$ & $R^2$ $\downarrow$ & $U$ $\downarrow$ & $U_0$ $\downarrow$ & ZPVE $\downarrow$\\
\midrule

$\lambda_2=0$ & 0.057 & 43.15 & 26.05 & 21.42 & 0.027 & 0.030 & 12.23 & 11.95 & 0.162 & 12.20 & 11.42 & 1.594\\

$\lambda_2=0.01$ & 0.056 & 42.73 & 25.95 & 21.55 & 0.027 & 0.029 & 11.22 & 10.70 & 0.141 & 10.81 & 10.70 & 1.559\\

$\lambda_2=1$ & 0.066 & 45.45 & 28.23 & 23.67 & 0.028 & 0.030 & 14.67 & 14.42 & 0.204 & 13.30 & 13.25 & 1.797\\

\bottomrule
\end{tabular}
\end{adjustbox}
\vspace{-2ex}
\end{table}

\begin{table}[tb]
\setlength{\tabcolsep}{5pt}
\fontsize{9}{9}\selectfont
\centering
\caption{
\small
Ablation studies of contrastive loss term in \model{}. The ablation results are on MD17.
}
\label{tab:ablation_contrastive_loss_MD17_result}
\vspace{-1.5ex}
\begin{adjustbox}{max width=\textwidth}
\begin{tabular}{l c c c c c c c c}
\toprule
Pretraining & Aspirin $\downarrow$ & Benzene $\downarrow$ & Ethanol $\downarrow$ & Malonaldehyde $\downarrow$ & Naphthalene $\downarrow$ & Salicylic $\downarrow$ & Toluene $\downarrow$ & Uracil $\downarrow$ \\
\midrule

$\lambda_2=0$ & 1.380 & 0.359 & 0.363 & 0.744 & 0.482 & 0.902 & 0.548 & 0.590\\

$\lambda_2=0.01$ & 1.289 & 0.345 & 0.365 & 0.613 & 0.498 & 0.712 & 0.480 & 0.463\\

$\lambda_2=1$ & 1.561 & 0.547 & 0.781 & 0.735 & 0.918 & 1.160 & 1.052 & 0.809\\

\bottomrule
\end{tabular}
\end{adjustbox}
\vspace{-3ex}
\end{table}

\subsection{Molecular Dynamics Prediction}
\vspace{-0.5ex}
MD17~\cite{chmiela2017machine} is a dataset on molecular dynamics simulation. It contains eight tasks corresponding to eight organic molecules, and the goal is to predict the forces at different 3D positions. The size of each task ranges from 48K to 991K, and please check~\Cref{sec:dataset_specification} for details.
The main results are in~\Cref{tab:main_MD17_result}, and \model{} can outperform 12 baselines on 6 out of 8 tasks and reach the second-best for one of the remaining tasks.

\subsection{Ablation Study on the Effectiveness of Contrastive Loss}
\vspace{-0.5ex}
As discussed in~\Cref{eq:final_objective}, there is one important hyperparameter $\lambda_2$ controlling the contrastive loss in \model{}. We want to conduct an ablation study on the effect of this contrastive term. As shown in~\Cref{tab:ablation_contrastive_loss_QM9_result,tab:ablation_contrastive_loss_MD17_result}, we consider three value for $\lambda_2$: 0, 0.01, and 1. $\lambda_2=0$ simply means that we only consider the reconstructive task, and its performance is very close to $\lambda_2=0.01$, {\ie}, the optimal results reported in~\Cref{tab:main_QM9_result,tab:main_MD17_result}. However, as we increase the $\lambda_2=1$, the performance degrades by a large margin. Thus, we would like to claim that the contrastive term in \model{} is comparatively sensitive, and we need to tune this hyperparameter carefully to obtain optimal results.

\section{Conclusion}
\vspace{-1ex}
In this work, we introduce a novel joint self-supervised learning framework called \model{}, which is based on augmented trajectory modeling. The term ``joint'' in our framework has two implications: Firstly, it signifies that our method is designed to model the joint distribution of trajectories rather than solely focusing on the marginal distribution. Secondly, it indicates that the augmented molecule trajectory incorporates both 2D and 3D information, providing insights into different aspects of molecule representation. While our proposed method has demonstrated the best empirical results on 15 out of 20 geometry-related property prediction tasks, there are still areas left for improvement, such as architecture design. Please refer to Appendix \ref{sec:app:limit} for an in-depth discussion.

\section*{Acknowledgement}
Jiujiu Chen would like to thank for her Associate Researcher Ruidong Chen, and Professor Xiaosong Zhang in UESTC for their special support.

\bibliographystyle{unsrtnat}
\bibliography{main.bib}

\begin{thebibliography}{76}
\providecommand{\natexlab}[1]{#1}
\providecommand{\url}[1]{\texttt{#1}}
\expandafter\ifx\csname urlstyle\endcsname\relax
  \providecommand{\doi}[1]{doi: #1}\else
  \providecommand{\doi}{doi: \begingroup \urlstyle{rm}\Url}\fi

\bibitem[Hu et~al.(2019)Hu, Liu, Gomes, Zitnik, Liang, Pande, and
  Leskovec]{hu2019strategies}
Weihua Hu, Bowen Liu, Joseph Gomes, Marinka Zitnik, Percy Liang, Vijay Pande,
  and Jure Leskovec.
\newblock Strategies for pre-training graph neural networks.
\newblock \emph{arXiv preprint arXiv:1905.12265}, 2019.

\bibitem[Liu et~al.(2019)Liu, Demirel, and Liang]{liu2019n}
Shengchao Liu, Mehmet~F Demirel, and Yingyu Liang.
\newblock N-gram graph: Simple unsupervised representation for graphs, with
  applications to molecules.
\newblock \emph{Advances in neural information processing systems}, 32, 2019.

\bibitem[Hu et~al.(2020{\natexlab{a}})Hu, Dong, Wang, Chang, and
  Sun]{hu2020gpt}
Ziniu Hu, Yuxiao Dong, Kuansan Wang, Kai-Wei Chang, and Yizhou Sun.
\newblock Gpt-gnn: Generative pre-training of graph neural networks.
\newblock In \emph{Proceedings of the 26th ACM SIGKDD International Conference
  on Knowledge Discovery \& Data Mining}, pages 1857--1867, 2020{\natexlab{a}}.

\bibitem[Sun et~al.(2020)Sun, Hoffmann, Verma, and Tang]{sun2019infograph}
Fan-Yun Sun, Jordan Hoffmann, Vikas Verma, and Jian Tang.
\newblock Infograph: Unsupervised and semi-supervised graph-level
  representation learning via mutual information maximization.
\newblock In \emph{International Conference on Learning Representations, ICLR},
  2020.

\bibitem[Liu et~al.(2023{\natexlab{a}})Liu, Guo, and Tang]{liu2022molecular}
Shengchao Liu, Hongyu Guo, and Jian Tang.
\newblock Molecular geometry pretraining with {SE}(3)-invariant denoising
  distance matching.
\newblock In \emph{The Eleventh International Conference on Learning
  Representations}, 2023{\natexlab{a}}.
\newblock URL \url{https://openreview.net/forum?id=CjTHVo1dvR}.

\bibitem[Zaidi et~al.(2022)Zaidi, Schaarschmidt, Martens, Kim, Teh,
  Sanchez-Gonzalez, Battaglia, Pascanu, and Godwin]{zaidi2022pre}
Sheheryar Zaidi, Michael Schaarschmidt, James Martens, Hyunjik Kim, Yee~Whye
  Teh, Alvaro Sanchez-Gonzalez, Peter Battaglia, Razvan Pascanu, and Jonathan
  Godwin.
\newblock Pre-training via denoising for molecular property prediction.
\newblock \emph{arXiv preprint arXiv:2206.00133}, 2022.

\bibitem[Liu et~al.(2021{\natexlab{a}})Liu, Wang, Liu, Lasenby, Guo, and
  Tang]{liu2021pre}
Shengchao Liu, Hanchen Wang, Weiyang Liu, Joan Lasenby, Hongyu Guo, and Jian
  Tang.
\newblock Pre-training molecular graph representation with 3d geometry.
\newblock \emph{arXiv preprint arXiv:2110.07728}, 2021{\natexlab{a}}.

\bibitem[Liu et~al.(2023{\natexlab{b}})Liu, Du, Ma, Guo, and
  Tang]{liu2023group}
Shengchao Liu, Weitao Du, Zhi-Ming Ma, Hongyu Guo, and Jian Tang.
\newblock A group symmetric stochastic differential equation model for molecule
  multi-modal pretraining.
\newblock In \emph{International Conference on Machine Learning}, pages
  21497--21526. PMLR, 2023{\natexlab{b}}.

\bibitem[Liu et~al.(2023{\natexlab{c}})Liu, Du, Li, Li, Zheng, Duan, Ma, Yaghi,
  Anandkumar, Borgs, et~al.]{liu2023symmetry}
Shengchao Liu, Weitao Du, Yanjing Li, Zhuoxinran Li, Zhiling Zheng, Chenru
  Duan, Zhiming Ma, Omar Yaghi, Anima Anandkumar, Christian Borgs, et~al.
\newblock Symmetry-informed geometric representation for molecules, proteins,
  and crystalline materials.
\newblock \emph{arXiv preprint arXiv:2306.09375}, 2023{\natexlab{c}}.

\bibitem[Liu et~al.(2021{\natexlab{b}})Liu, Pan, Jin, Zhou, Xia, and
  Yu]{liu2021graph}
Yixin Liu, Shirui Pan, Ming Jin, Chuan Zhou, Feng Xia, and Philip~S Yu.
\newblock Graph self-supervised learning: A survey.
\newblock \emph{arXiv preprint arXiv:2103.00111}, 2021{\natexlab{b}}.

\bibitem[Xie et~al.(2021)Xie, Xu, Zhang, Wang, and Ji]{xie2021self}
Yaochen Xie, Zhao Xu, Jingtun Zhang, Zhengyang Wang, and Shuiwang Ji.
\newblock Self-supervised learning of graph neural networks: A unified review.
\newblock \emph{arXiv preprint arXiv:2102.10757}, 2021.

\bibitem[Wu et~al.(2021)Wu, Lin, Gao, Tan, Li, et~al.]{wu2021self}
Lirong Wu, Haitao Lin, Zhangyang Gao, Cheng Tan, Stan Li, et~al.
\newblock Self-supervised on graphs: Contrastive, generative, or predictive.
\newblock \emph{arXiv preprint arXiv:2105.07342}, 2021.

\bibitem[Liu et~al.(2021{\natexlab{c}})Liu, Zhang, Hou, Mian, Wang, Zhang, and
  Tang]{liu2021self}
Xiao Liu, Fanjin Zhang, Zhenyu Hou, Li~Mian, Zhaoyu Wang, Jing Zhang, and Jie
  Tang.
\newblock Self-supervised learning: Generative or contrastive.
\newblock \emph{IEEE Transactions on Knowledge and Data Engineering},
  2021{\natexlab{c}}.

\bibitem[Madotto et~al.(2020)Madotto, Liu, Lin, and Fung]{madotto2020language}
Andrea Madotto, Zihan Liu, Zhaojiang Lin, and Pascale Fung.
\newblock Language models as few-shot learner for task-oriented dialogue
  systems.
\newblock \emph{arXiv preprint arXiv:2008.06239}, 2020.

\bibitem[Zhao et~al.(2023)Zhao, Zhou, Li, Tang, Wang, Hou, Min, Zhang, Zhang,
  Dong, et~al.]{zhao2023survey}
Wayne~Xin Zhao, Kun Zhou, Junyi Li, Tianyi Tang, Xiaolei Wang, Yupeng Hou,
  Yingqian Min, Beichen Zhang, Junjie Zhang, Zican Dong, et~al.
\newblock A survey of large language models.
\newblock \emph{arXiv preprint arXiv:2303.18223}, 2023.

\bibitem[Vahdat et~al.(2021)Vahdat, Kreis, and Kautz]{vahdat2021score}
Arash Vahdat, Karsten Kreis, and Jan Kautz.
\newblock Score-based generative modeling in latent space.
\newblock \emph{Advances in Neural Information Processing Systems},
  34:\penalty0 11287--11302, 2021.

\bibitem[Ramesh et~al.(2022)Ramesh, Dhariwal, Nichol, Chu, and
  Chen]{ramesh2022hierarchical}
Aditya Ramesh, Prafulla Dhariwal, Alex Nichol, Casey Chu, and Mark Chen.
\newblock Hierarchical text-conditional image generation with clip latents.
\newblock \emph{arXiv preprint arXiv:2204.06125}, 2022.

\bibitem[Thomas et~al.(2018)Thomas, Smidt, Kearnes, Yang, Li, Kohlhoff, and
  Riley]{thomas2018tensor}
Nathaniel Thomas, Tess Smidt, Steven Kearnes, Lusann Yang, Li~Li, Kai Kohlhoff,
  and Patrick Riley.
\newblock Tensor field networks: Rotation-and translation-equivariant neural
  networks for 3d point clouds.
\newblock \emph{arXiv preprint arXiv:1802.08219}, 2018.

\bibitem[Wood and Erpenbeck(1976)]{wood1976molecular}
William~W Wood and Jerome~J Erpenbeck.
\newblock Molecular dynamics and monte carlo calculations in statistical
  mechanics.
\newblock \emph{Annual Review of Physical Chemistry}, 27\penalty0 (1):\penalty0
  319--348, 1976.

\bibitem[Sohl-Dickstein et~al.(2015)Sohl-Dickstein, Weiss, Maheswaranathan, and
  Ganguli]{sohl2015deep}
Jascha Sohl-Dickstein, Eric Weiss, Niru Maheswaranathan, and Surya Ganguli.
\newblock Deep unsupervised learning using nonequilibrium thermodynamics.
\newblock In \emph{International Conference on Machine Learning}, pages
  2256--2265. PMLR, 2015.

\bibitem[Dockhorn et~al.(2021)Dockhorn, Vahdat, and Kreis]{dockhorn2021score}
Tim Dockhorn, Arash Vahdat, and Karsten Kreis.
\newblock Score-based generative modeling with critically-damped langevin
  diffusion.
\newblock \emph{arXiv preprint arXiv:2112.07068}, 2021.

\bibitem[Jo et~al.(2022)Jo, Lee, and Hwang]{jo2022score}
Jaehyeong Jo, Seul Lee, and Sung~Ju Hwang.
\newblock Score-based generative modeling of graphs via the system of
  stochastic differential equations.
\newblock In \emph{International Conference on Machine Learning}, pages
  10362--10383. PMLR, 2022.

\bibitem[Vignac et~al.(2022)Vignac, Krawczuk, Siraudin, Wang, Cevher, and
  Frossard]{vignac2022digress}
Clement Vignac, Igor Krawczuk, Antoine Siraudin, Bohan Wang, Volkan Cevher, and
  Pascal Frossard.
\newblock Digress: Discrete denoising diffusion for graph generation.
\newblock \emph{arXiv preprint arXiv:2209.14734}, 2022.

\bibitem[Du et~al.(2022{\natexlab{a}})Du, Zhang, Du, Meng, Chen, Zheng, Shao,
  and Liu]{du2022se}
Weitao Du, He~Zhang, Yuanqi Du, Qi~Meng, Wei Chen, Nanning Zheng, Bin Shao, and
  Tie-Yan Liu.
\newblock Se (3) equivariant graph neural networks with complete local frames.
\newblock In \emph{International Conference on Machine Learning}, pages
  5583--5608. PMLR, 2022{\natexlab{a}}.

\bibitem[Liu et~al.(2023{\natexlab{d}})Liu, Du, Ma, Guo, and Tang]{liugroup}
Shengchao Liu, Weitao Du, Zhiming Ma, Hongyu Guo, and Jian Tang.
\newblock A group symmetric stochastic differential equation model for molecule
  multi-modal pretraining.
\newblock 2023{\natexlab{d}}.

\bibitem[Shi et~al.(2021)Shi, Luo, Xu, and Tang]{shi2021learning}
Chence Shi, Shitong Luo, Minkai Xu, and Jian Tang.
\newblock Learning gradient fields for molecular conformation generation.
\newblock In \emph{International Conference on Machine Learning}, pages
  9558--9568. PMLR, 2021.

\bibitem[Hoogeboom et~al.(2022)Hoogeboom, Satorras, Vignac, and
  Welling]{hoogeboom2022equivariant}
Emiel Hoogeboom, V{\i}ctor~Garcia Satorras, Cl{\'e}ment Vignac, and Max
  Welling.
\newblock Equivariant diffusion for molecule generation in 3d.
\newblock In \emph{International Conference on Machine Learning}, pages
  8867--8887. PMLR, 2022.

\bibitem[Song et~al.(2020)Song, Sohl-Dickstein, Kingma, Kumar, Ermon, and
  Poole]{song2020score}
Yang Song, Jascha Sohl-Dickstein, Diederik~P Kingma, Abhishek Kumar, Stefano
  Ermon, and Ben Poole.
\newblock Score-based generative modeling through stochastic differential
  equations.
\newblock \emph{arXiv preprint arXiv:2011.13456}, 2020.

\bibitem[Du et~al.(2022{\natexlab{b}})Du, Yang, Zhang, and Du]{du2022flexible}
Weitao Du, Tao Yang, He~Zhang, and Yuanqi Du.
\newblock A flexible diffusion model.
\newblock \emph{arXiv preprint arXiv:2206.10365}, 2022{\natexlab{b}}.

\bibitem[Austin et~al.(2021)Austin, Johnson, Ho, Tarlow, and van~den
  Berg]{austin2021structured}
Jacob Austin, Daniel~D Johnson, Jonathan Ho, Daniel Tarlow, and Rianne van~den
  Berg.
\newblock Structured denoising diffusion models in discrete state-spaces.
\newblock \emph{Advances in Neural Information Processing Systems},
  34:\penalty0 17981--17993, 2021.

\bibitem[Li et~al.(2022)Li, Thickstun, Gulrajani, Liang, and
  Hashimoto]{li2022diffusion}
Xiang Li, John Thickstun, Ishaan Gulrajani, Percy~S Liang, and Tatsunori~B
  Hashimoto.
\newblock Diffusion-lm improves controllable text generation.
\newblock \emph{Advances in Neural Information Processing Systems},
  35:\penalty0 4328--4343, 2022.

\bibitem[Ho et~al.(2020)Ho, Jain, and Abbeel]{ho2020denoising}
Jonathan Ho, Ajay Jain, and Pieter Abbeel.
\newblock Denoising diffusion probabilistic models.
\newblock \emph{Advances in Neural Information Processing Systems},
  33:\penalty0 6840--6851, 2020.

\bibitem[Bansal et~al.(2022)Bansal, Borgnia, Chu, Li, Kazemi, Huang, Goldblum,
  Geiping, and Goldstein]{bansal2022cold}
Arpit Bansal, Eitan Borgnia, Hong-Min Chu, Jie~S Li, Hamid Kazemi, Furong
  Huang, Micah Goldblum, Jonas Geiping, and Tom Goldstein.
\newblock Cold diffusion: Inverting arbitrary image transforms without noise.
\newblock \emph{arXiv preprint arXiv:2208.09392}, 2022.

\bibitem[Goodfellow et~al.(2020)Goodfellow, Pouget-Abadie, Mirza, Xu,
  Warde-Farley, Ozair, Courville, and Bengio]{goodfellow2020generative}
Ian Goodfellow, Jean Pouget-Abadie, Mehdi Mirza, Bing Xu, David Warde-Farley,
  Sherjil Ozair, Aaron Courville, and Yoshua Bengio.
\newblock Generative adversarial networks.
\newblock \emph{Communications of the ACM}, 63\penalty0 (11):\penalty0
  139--144, 2020.

\bibitem[Kingma and Welling(2013)]{kingma2013auto}
Diederik~P Kingma and Max Welling.
\newblock Auto-encoding variational bayes.
\newblock \emph{arXiv preprint arXiv:1312.6114}, 2013.

\bibitem[Tian et~al.(2020)Tian, Sun, Poole, Krishnan, Schmid, and
  Isola]{tian2020makes}
Yonglong Tian, Chen Sun, Ben Poole, Dilip Krishnan, Cordelia Schmid, and
  Phillip Isola.
\newblock What makes for good views for contrastive learning?
\newblock \emph{Advances in neural information processing systems},
  33:\penalty0 6827--6839, 2020.

\bibitem[Sinha et~al.(2021)Sinha, Ayush, Song, Uzkent, Jin, and
  Ermon]{sinha2021negative}
Abhishek Sinha, Kumar Ayush, Jiaming Song, Burak Uzkent, Hongxia Jin, and
  Stefano Ermon.
\newblock Negative data augmentation.
\newblock \emph{arXiv preprint arXiv:2102.05113}, 2021.

\bibitem[Nichol and Dhariwal(2021)]{nichol2021improved}
Alexander~Quinn Nichol and Prafulla Dhariwal.
\newblock Improved denoising diffusion probabilistic models.
\newblock In \emph{International Conference on Machine Learning}, pages
  8162--8171. PMLR, 2021.

\bibitem[Preechakul et~al.(2022)Preechakul, Chatthee, Wizadwongsa, and
  Suwajanakorn]{preechakul2022diffusion}
Konpat Preechakul, Nattanat Chatthee, Suttisak Wizadwongsa, and Supasorn
  Suwajanakorn.
\newblock Diffusion autoencoders: Toward a meaningful and decodable
  representation.
\newblock In \emph{Proceedings of the IEEE/CVF Conference on Computer Vision
  and Pattern Recognition}, pages 10619--10629, 2022.

\bibitem[Zhang and Agrawala(2023)]{zhang2023adding}
Lvmin Zhang and Maneesh Agrawala.
\newblock Adding conditional control to text-to-image diffusion models.
\newblock \emph{arXiv preprint arXiv:2302.05543}, 2023.

\bibitem[Yang et~al.(2022)Yang, Xiao, Zhang, Guo, Zhao, and
  Shen]{yang2022image}
Suorong Yang, Weikang Xiao, Mengcheng Zhang, Suhan Guo, Jian Zhao, and Furao
  Shen.
\newblock Image data augmentation for deep learning: A survey.
\newblock \emph{arXiv preprint arXiv:2204.08610}, 2022.

\bibitem[Trabucco et~al.(2023)Trabucco, Doherty, Gurinas, and
  Salakhutdinov]{trabucco2023effective}
Brandon Trabucco, Kyle Doherty, Max Gurinas, and Ruslan Salakhutdinov.
\newblock Effective data augmentation with diffusion models.
\newblock \emph{arXiv preprint arXiv:2302.07944}, 2023.

\bibitem[Zhang et~al.(2022{\natexlab{a}})Zhang, Tao, and Chen]{zhang2022gddim}
Qinsheng Zhang, Molei Tao, and Yongxin Chen.
\newblock gddim: Generalized denoising diffusion implicit models.
\newblock \emph{arXiv preprint arXiv:2206.05564}, 2022{\natexlab{a}}.

\bibitem[Huang et~al.(2021)Huang, Lim, and Courville]{huang2021variational}
Chin-Wei Huang, Jae~Hyun Lim, and Aaron~C Courville.
\newblock A variational perspective on diffusion-based generative models and
  score matching.
\newblock \emph{Advances in Neural Information Processing Systems},
  34:\penalty0 22863--22876, 2021.

\bibitem[Satorras et~al.(2021)Satorras, Hoogeboom, and Welling]{satorras2021n}
V{\i}ctor~Garcia Satorras, Emiel Hoogeboom, and Max Welling.
\newblock E (n) equivariant graph neural networks.
\newblock In \emph{International conference on machine learning}, pages
  9323--9332. PMLR, 2021.

\bibitem[Chmiela et~al.(2017)Chmiela, Tkatchenko, Sauceda, Poltavsky,
  Sch{\"u}tt, and M{\"u}ller]{chmiela2017machine}
Stefan Chmiela, Alexandre Tkatchenko, Huziel~E Sauceda, Igor Poltavsky,
  Kristof~T Sch{\"u}tt, and Klaus-Robert M{\"u}ller.
\newblock Machine learning of accurate energy-conserving molecular force
  fields.
\newblock \emph{Science advances}, 3\penalty0 (5):\penalty0 e1603015, 2017.

\bibitem[Goan and Fookes(2020)]{goan2020bayesian}
Ethan Goan and Clinton Fookes.
\newblock Bayesian neural networks: An introduction and survey.
\newblock \emph{Case Studies in Applied Bayesian Data Science: CIRM Jean-Morlet
  Chair, Fall 2018}, pages 45--87, 2020.

\bibitem[Grathwohl et~al.(2019)Grathwohl, Wang, Jacobsen, Duvenaud, Norouzi,
  and Swersky]{grathwohl2019your}
Will Grathwohl, Kuan-Chieh Wang, J{\"o}rn-Henrik Jacobsen, David Duvenaud,
  Mohammad Norouzi, and Kevin Swersky.
\newblock Your classifier is secretly an energy based model and you should
  treat it like one.
\newblock \emph{arXiv preprint arXiv:1912.03263}, 2019.

\bibitem[Zhao et~al.(2020)Zhao, Jacobsen, and Grathwohl]{zhao2020joint}
Stephen Zhao, Jorn-Henrik Jacobsen, and Will Grathwohl.
\newblock Joint energy-based models for semi-supervised classification.
\newblock In \emph{ICML 2020 Workshop on Uncertainty and Robustness in Deep
  Learning}, volume~1, 2020.

\bibitem[Li and Kluger(2022)]{li2022autoregressive}
Henry Li and Yuval Kluger.
\newblock Autoregressive generative modeling with noise conditional maximum
  likelihood estimation.
\newblock \emph{arXiv preprint arXiv:2210.10715}, 2022.

\bibitem[Chen et~al.(2020)Chen, Kornblith, Norouzi, and Hinton]{chen2020simple}
Ting Chen, Simon Kornblith, Mohammad Norouzi, and Geoffrey Hinton.
\newblock A simple framework for contrastive learning of visual
  representations.
\newblock In \emph{International conference on machine learning}, pages
  1597--1607. PMLR, 2020.

\bibitem[Gupta et~al.(2022)Gupta, Ajanthan, Hengel, and
  Gould]{gupta2022understanding}
Kartik Gupta, Thalaiyasingam Ajanthan, Anton van~den Hengel, and Stephen Gould.
\newblock Understanding and improving the role of projection head in
  self-supervised learning.
\newblock \emph{arXiv preprint arXiv:2212.11491}, 2022.

\bibitem[Meng et~al.(2021)Meng, He, Song, Song, Wu, Zhu, and
  Ermon]{meng2021sdedit}
Chenlin Meng, Yutong He, Yang Song, Jiaming Song, Jiajun Wu, Jun-Yan Zhu, and
  Stefano Ermon.
\newblock Sdedit: Guided image synthesis and editing with stochastic
  differential equations.
\newblock In \emph{International Conference on Learning Representations}, 2021.

\bibitem[Du et~al.(2023)Du, Du, Wang, Feng, Wang, Ji, Gomes, and Ma]{du2023new}
Weitao Du, Yuanqi Du, Limei Wang, Dieqiao Feng, Guifeng Wang, Shuiwang Ji,
  Carla Gomes, and Zhi-Ming Ma.
\newblock A new perspective on building efficient and expressive 3d equivariant
  graph neural networks.
\newblock \emph{arXiv preprint arXiv:2304.04757}, 2023.

\bibitem[Hu et~al.(2020{\natexlab{b}})Hu, Fey, Zitnik, Dong, Ren, Liu, Catasta,
  and Leskovec]{hu2020ogb}
Weihua Hu, Matthias Fey, Marinka Zitnik, Yuxiao Dong, Hongyu Ren, Bowen Liu,
  Michele Catasta, and Jure Leskovec.
\newblock Open graph benchmark: Datasets for machine learning on graphs.
\newblock \emph{arXiv preprint arXiv:2005.00687}, 2020{\natexlab{b}}.

\bibitem[Nakata and Shimazaki(2017)]{nakata2017pubchemqc}
Maho Nakata and Tomomi Shimazaki.
\newblock Pubchemqc project: a large-scale first-principles electronic
  structure database for data-driven chemistry.
\newblock \emph{Journal of chemical information and modeling}, 57\penalty0
  (6):\penalty0 1300--1308, 2017.

\bibitem[Sch{\"u}tt et~al.(2018)Sch{\"u}tt, Sauceda, Kindermans, Tkatchenko,
  and M{\"u}ller]{schutt2018schnet}
Kristof~T Sch{\"u}tt, Huziel~E Sauceda, P-J Kindermans, Alexandre Tkatchenko,
  and K-R M{\"u}ller.
\newblock Schnet--a deep learning architecture for molecules and materials.
\newblock \emph{The Journal of Chemical Physics}, 148\penalty0 (24):\penalty0
  241722, 2018.

\bibitem[St{\"a}rk et~al.(2022)St{\"a}rk, Beaini, Corso, Tossou, Dallago,
  G{\"u}nnemann, and Li{\`o}]{stark20223d}
Hannes St{\"a}rk, Dominique Beaini, Gabriele Corso, Prudencio Tossou, Christian
  Dallago, Stephan G{\"u}nnemann, and Pietro Li{\`o}.
\newblock 3d infomax improves gnns for molecular property prediction.
\newblock In \emph{International Conference on Machine Learning}, pages
  20479--20502. PMLR, 2022.

\bibitem[Ramakrishnan et~al.(2014)Ramakrishnan, Dral, Rupp, and
  Von~Lilienfeld]{ramakrishnan2014quantum}
Raghunathan Ramakrishnan, Pavlo~O Dral, Matthias Rupp, and O~Anatole
  Von~Lilienfeld.
\newblock Quantum chemistry structures and properties of 134 kilo molecules.
\newblock \emph{Scientific data}, 1\penalty0 (1):\penalty0 1--7, 2014.

\bibitem[Jing et~al.(2022)Jing, Corso, Chang, Barzilay, and
  Jaakkola]{jing2022torsional}
Bowen Jing, Gabriele Corso, Jeffrey Chang, Regina Barzilay, and Tommi Jaakkola.
\newblock Torsional diffusion for molecular conformer generation.
\newblock \emph{arXiv preprint arXiv:2206.01729}, 2022.

\bibitem[Baek et~al.(2021)Baek, Kang, and Hwang]{baek2021accurate}
Jinheon Baek, Minki Kang, and Sung~Ju Hwang.
\newblock Accurate learning of graph representations with graph multiset
  pooling.
\newblock \emph{arXiv preprint arXiv:2102.11533}, 2021.

\bibitem[Huang et~al.(2022)Huang, Zhang, Xu, and Wong]{huang2022mdm}
Lei Huang, Hengtong Zhang, Tingyang Xu, and Ka-Chun Wong.
\newblock Mdm: Molecular diffusion model for 3d molecule generation, 2022.

\bibitem[Halgren(1996)]{halgren1996merck}
Thomas~A Halgren.
\newblock Merck molecular force field. i. basis, form, scope, parameterization,
  and performance of mmff94.
\newblock \emph{Journal of computational chemistry}, 17\penalty0
  (5-6):\penalty0 490--519, 1996.

\bibitem[Balestriero and LeCun(2022)]{balestriero2022contrastive}
Randall Balestriero and Yann LeCun.
\newblock Contrastive and non-contrastive self-supervised learning recover
  global and local spectral embedding methods.
\newblock \emph{arXiv preprint arXiv:2205.11508}, 2022.

\bibitem[Mathieu et~al.(2016)Mathieu, Zhao, Zhao, Ramesh, Sprechmann, and
  LeCun]{mathieu2016disentangling}
Michael~F Mathieu, Junbo~Jake Zhao, Junbo Zhao, Aditya Ramesh, Pablo
  Sprechmann, and Yann LeCun.
\newblock Disentangling factors of variation in deep representation using
  adversarial training.
\newblock \emph{Advances in neural information processing systems}, 29, 2016.

\bibitem[Song et~al.(2023)Song, Dhariwal, Chen, and
  Sutskever]{song2023consistency}
Yang Song, Prafulla Dhariwal, Mark Chen, and Ilya Sutskever.
\newblock Consistency models.
\newblock \emph{arXiv preprint arXiv:2303.01469}, 2023.

\bibitem[Zhu et~al.(2021)Zhu, Sun, and Li]{zhu2021rethinking}
Yao Zhu, Jiacheng Sun, and Zhenguo Li.
\newblock Rethinking adversarial transferability from a data distribution
  perspective.
\newblock In \emph{International Conference on Learning Representations}, 2021.

\bibitem[Zhang et~al.(2022{\natexlab{b}})Zhang, Wang, and Wang]{zhang2022mask}
Qi~Zhang, Yifei Wang, and Yisen Wang.
\newblock How mask matters: Towards theoretical understandings of masked
  autoencoders.
\newblock \emph{arXiv preprint arXiv:2210.08344}, 2022{\natexlab{b}}.

\bibitem[Wang et~al.(2022)Wang, Wang, Cao, and
  Barati~Farimani]{wang2022molecular}
Yuyang Wang, Jianren Wang, Zhonglin Cao, and Amir Barati~Farimani.
\newblock Molecular contrastive learning of representations via graph neural
  networks.
\newblock \emph{Nature Machine Intelligence}, 4\penalty0 (3):\penalty0
  279--287, 2022.

\bibitem[Liu et~al.(2023{\natexlab{e}})Liu, Guo, and Tang]{liu2023molecular}
Shengchao Liu, Hongyu Guo, and Jian Tang.
\newblock Molecular geometry pretraining with {SE}(3)-invariant denoising
  distance matching.
\newblock In \emph{The Eleventh International Conference on Learning
  Representations}, 2023{\natexlab{e}}.
\newblock URL \url{https://openreview.net/forum?id=CjTHVo1dvR}.

\bibitem[Zhou et~al.(2023)Zhou, Gao, Ding, Zheng, Xu, Wei, Zhang, and
  Ke]{zhou2023uni}
Gengmo Zhou, Zhifeng Gao, Qiankun Ding, Hang Zheng, Hongteng Xu, Zhewei Wei,
  Linfeng Zhang, and Guolin Ke.
\newblock Uni-mol: A universal 3d molecular representation learning framework.
\newblock 2023.

\bibitem[Daras et~al.(2022)Daras, Delbracio, Talebi, Dimakis, and
  Milanfar]{daras2022soft}
Giannis Daras, Mauricio Delbracio, Hossein Talebi, Alexandros~G Dimakis, and
  Peyman Milanfar.
\newblock Soft diffusion: Score matching for general corruptions.
\newblock \emph{arXiv preprint arXiv:2209.05442}, 2022.

\bibitem[Lemley et~al.(2017)Lemley, Bazrafkan, and Corcoran]{lemley2017smart}
Joseph Lemley, Shabab Bazrafkan, and Peter Corcoran.
\newblock Smart augmentation learning an optimal data augmentation strategy.
\newblock \emph{Ieee Access}, 5:\penalty0 5858--5869, 2017.

\bibitem[Cubuk et~al.(2019)Cubuk, Zoph, Mane, Vasudevan, and
  Le]{cubuk2019autoaugment}
Ekin~D Cubuk, Barret Zoph, Dandelion Mane, Vijay Vasudevan, and Quoc~V Le.
\newblock Autoaugment: Learning augmentation strategies from data.
\newblock In \emph{Proceedings of the IEEE/CVF conference on computer vision
  and pattern recognition}, pages 113--123, 2019.

\bibitem[Vignac et~al.(2023)Vignac, Osman, Toni, and Frossard]{vignac2023midi}
Clement Vignac, Nagham Osman, Laura Toni, and Pascal Frossard.
\newblock Midi: Mixed graph and 3d denoising diffusion for molecule generation,
  2023.

\bibitem[Axelrod and G{\'o}mez-Bombarelli(2022)]{axelrod2022geom}
Simon Axelrod and Rafael G{\'o}mez-Bombarelli.
\newblock Geom, energy-annotated molecular conformations for property
  prediction and molecular generation.
\newblock \emph{Scientific Data}, 9\penalty0 (1):\penalty0 185, 2022.
\newblock \doi{10.1038/s41597-022-01288-4}.
\newblock URL \url{https://doi.org/10.1038/s41597-022-01288-4}.

\end{thebibliography}

\clearpage
\appendix

\section{Methodology Detail}
\subsection{Details for Section \ref{sec:reconstructive_and_contrastive_loss}}
Following Eq. \ref{marginal}, we consider the 'marginal' distribution 
$$q_{\theta}(x_0): = \int p_{\theta}(x_0,x_t)dx_t$$ by marginalizing out $x_t$.  We define the corresponding energy function with respect to $x_0$ as 
$$\Bar{\textbf{E}}_{\theta} (x_0) = - \log  \int \exp (-\textbf{E}_{\theta}(x_0, x_t)) d x_t .$$

From the definition of $\Bar{\textbf{E}}_{\theta} (x_0)$,  the distribution  $q_{\theta}(x_0)$ shares the same normalization constant with $q_{\theta}(x_0,x_t)$.
Therefore we have
$$\frac{\partial \log f_{\theta}(x_0,x_t)}{\partial \theta} = - \frac{\partial \Bar{\textbf{E}}_{\theta}(x_0, x_t)}{\partial \theta} + \frac{\partial \Bar{\textbf{E}}_{\theta} (x_0)}{\partial \theta},$$
where $f_{\theta}(x_0 , x_t)$ denotes the normalized conditional probability $q_{\theta}(x_t | x_0)$.  By the conditional probability formula, it is obvious that:
it's obvious that
\begin{equation} 
\frac{\partial \log q_{\theta}(x_0, x_t)}{\partial \theta} =  \frac{\partial \log f_{\theta}(x_0 , x_t)}{\partial \theta} +  \frac{\partial \log q_{\theta}(x_0)}{\partial \theta}.\end{equation}
By taking the empirical expectation with respect to the sampled pair  $(x_0,x_t) \sim p(x_0,x_t)$, we find that the gradient of the Maximum Likelihood allows the following decomposition:
\begin{equation} \label{decomposition} \small
\Tilde{\mathbb{E}}_{p(x_0,x_t)}\left[\frac{\partial \log p_{\theta}(x_0, x_t)}{\partial \theta} \right]=  \underbrace{\Tilde{\mathbb{E}}_{p(x_0)}\left[\frac{\partial \log q_{\theta}(x_0)}{\partial \theta}\right]}_{\textbf{Gradient of reconstruction}} + \underbrace{\Tilde{\mathbb{E}}_{p(x_0,x_t)}\left[\frac{\partial \log f_{\theta}(x_0 , x_t)}{\partial \theta}\right]}_{\textbf{Gradient of contrastive learning}} .\end{equation}
Here, we use  $\Tilde{\mathbb{E}}$ to denote the expectation with respect to the empirical expectation. In this way, we find that maximizing the likelihood $p_{\theta}(x_0,x_t)$ is equivalent to solving:
$$\text{argmax}_{\theta} \left\{\Tilde{\mathbb{E}}_{p(x_0)}\left[\frac{\partial \log q_{\theta}(x_0)}{\partial \theta}\right] + \Tilde{\mathbb{E}}_{p(x_0,x_t)}\left[\frac{\partial \log f_{\theta}(x_0 , x_t)}{\partial \theta}\right] \right\}.$$
\begin{figure}[!t]
    \centering
    \includegraphics[scale=0.6]{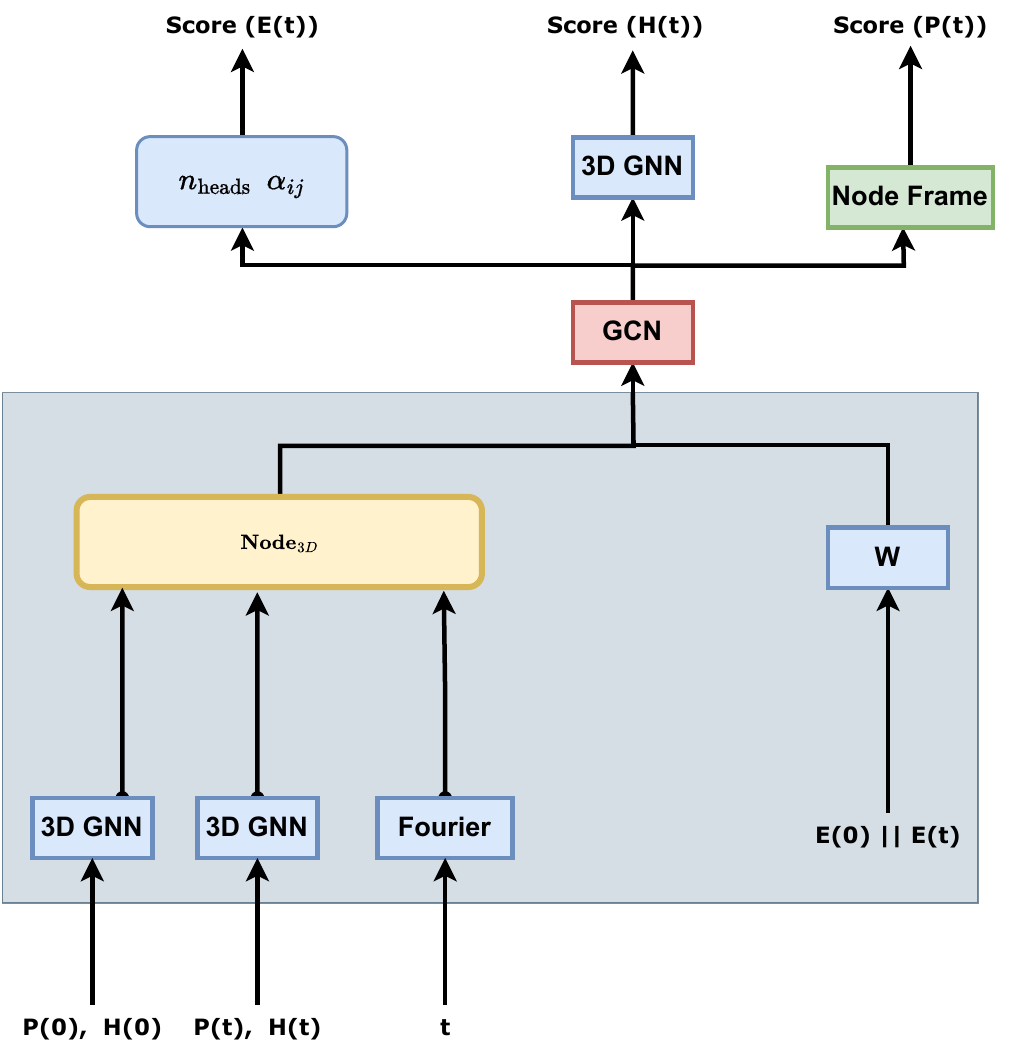}

    \caption{\small Modular Design of \model{}. }
    \label{fig:network}
    
\end{figure}
\subsection{Details for Section \ref{sec:model_architecture}}
In this section, we provide the details for the model design described in Section \ref{sec:model_architecture}. Specifically, we illustrate how to obtain the latent representation $h_{\theta}$ from $x_0 = (P_0,H_0,E_0)$ and $x_t = (P_t,H_t,E_t)$ (the solution of Eq. \ref{joint sde} at time $t$). It is important to note that in standard diffusion generative models \cite{ho2020denoising,song2020score}, only $x_t$ is fed into the score neural network.

The 3D part $(H_0, E_0)$ and $(H_t,E_t)$are transformed by two equivariant 3D GNNs, and we denote the invariant node-wise output representation by $f_0$ and $f_t$. Additionally, we embed the time $t$ (which is essential for trajectory learning) using Fourier embedding, following \cite{song2020score}:
$$\text{Emd}(t) : = \textbf{Fourier}(t).$$
We obtain the 3D node-wise representation by concatenating the Fourier embedding, $f_0$, and $f_t$:
$$\textbf{Node}_{3D} = \textbf{MLP}\ [\text{Emd}(t) \mathbin\Vert f_0 \mathbin\Vert f_t ].$$
Up to this point, we haven't utilized the 2D information $(E_0, E_t)$. Unlike the node-wise 3D representation, we encode $(E_0, E_t)$ as a weighted adjacency matrix $W$:
$$W = \textbf{MLP}\ [\text{Emd}(t) \mathbin\Vert E_0 \mathbin\Vert E_t ].$$
Combing $\textbf{Node}_{3D}$ and the adjacency matrix $W$, we implement a dense graph convolution neural network (GCN) to obtain the final (node-wise) latent representation  $h_{\theta}$, following \cite{jo2022score}:
$$h_{\theta} = \textbf{GCN}\ (\text{Node}_{3D}, W).$$
\paragraph{3D output $\text{Score}(P_t)$.} 
To obtain the $SE(3)$ equivariant and reflection anti-equivariant vector field \cite{jing2022torsional} $\text{Score}(P_t)$ from $h_{\theta}$, we implement the node-wise equivariant frame and perform tensorization, following \cite{du2022se,du2023new}. Consider node $i$ with 3D position $\textbf{x}_i$, let $\Bar{\textbf{x}}_i : = \frac{1}{N}\sum_{\textbf{x}_j \in \mathcal{N}(\textbf{x}_i)} \textbf{x}_j$ 
be the center of mass around $\textbf{x}_i$'s neighborhood. Then the orthonormal equivariant frame $\mathcal{F}_i : = (\textbf{e}^i_1,\textbf{e}^i_2,\textbf{e}^i_3)$ with respect to $\textbf{x}_i$ is defined as: 

\begin{equation} 
\label{eq:node frame}
(\frac{\textbf{x}_i - \Bar{\textbf{x}}_i}{\norm{\textbf{x}_i - \Bar{\textbf{x}}_i}}, \frac{\Bar{\textbf{x}}_i \times \textbf{x}_i}{\norm{\Bar{\textbf{x}}_i \times \textbf{x}_i}},\frac{\textbf{x}_i - \Bar{\textbf{x}}_i}{\norm{\textbf{x}_i - \Bar{\textbf{x}}_i}} \times \frac{\Bar{\textbf{x}}_i \times \textbf{x}_i}{\norm{\Bar{\textbf{x}}_i \times \textbf{x}_i}}  ),
\end{equation}
where $\times$ denotes the cross product, which is $SE(3)$ equivariant and reflection anti-equivariant. Then, we transform $h_{\theta} \in \mathbf{R}^{N \times L}$ to $h_{3D} \in \mathbf{R}^{N \times 3}$:
$$h_{3D} = (h_1,h_2,h_3) := \textbf{MLP}\ (h_{\theta}).$$
Finally, for node $i$,
$$\text{Score}^i(P_t) = h_1 \cdot \textbf{e}^i_1 + h_2 \cdot \textbf{e}^i_2 + h_3 \cdot \textbf{e}^i_3.$$
\paragraph{2D output $\text{Score}(E_t)$.} $\text{Score}(E_t)$ represents the probability gradient with respect to the molecule bonds. Therefore, $\text{Score}(E_t)$ has the same shape as the dense adjacency matrix and is $SE(3)$ invariant. To obtain $\text{Score}(E_t)$, we leverage graph multi-head attention \cite{baek2021accurate} to obtain a dense attention matrix $A_t$ (see \cite{baek2021accurate} for the explicit formula):
$$A_t = \textbf{Att}\ (h_{\theta}).$$
Then, we apply an \textbf{MLP} to the edge-wise $A_t$ to obtain the final 2D score function:
$$\text{Score}(E_t) = \textbf{MLP}\ (A_t).$$
See Figure \ref{fig:network} for a graphical representation. 
\subsection{A Cold Alternative for Section \ref{sec:trajectory_construction}}

Unlike continuous data that can take values ranging from $(-\infty, \infty)$, categorical data only takes a finite number of values.  For example, in this paper, the atom type takes its value from the periodic table of elements, while the bond type takes its value from the set $\{0, 1, 2, 3, 4\}$. For these finite state spaces, we can replace our continuous diffusion framework with a discrete (cold) diffusion framework.

To define a discrete diffusion, we need to specify the transition matrix $Q_t$ for each time $t$, as stated in Eq. (\ref{absorbing}). Following \cite{vignac2022digress,du2022flexible,huang2022mdm}, to effectively estimate the likelihood of the diffusion model, we further require that the noised state $q(z_t | x)$, defined as:
$$q(z_t | x) : = x  Q_1\dots Q_t$$
to be equivariant under permutation. This requirement can be satisfied if we diffuse
\textbf{separately on each node and edge feature}. Additionally, we require that the limiting distribution as  $t \rightarrow \infty$ should not depend on the original data $x$, so that we can use it as a prior distribution for inference. A common choice is a uniform transition over the classes, where the corresponding  $Q_t$ is defined as: $$Q_t : = \alpha_t \textbf{I} + (1- \alpha_t)\textbf{1}_d \textbf{1}_d^T/d$$ with $\alpha_t$ transitioning from 1 to 0. 
By letting $Q_t$ act on each node and each edge, we have defined the discrete diffusion for  $H(t)$ and $E(t)$:
$$q(H_t | H_{t-1}) = H_{t-1} \cdot Q_t,\ \ \ \text{and}\ \ \  q(E_t | E_{t-1}) = E_{t-1} \cdot Q_t.$$
On the other hand, although the 3D structure is represented as a continuous point cloud, the molecule's conformer structures at the stationary states are discrete \cite{liu2022molecular} and labeled by the energy levels of molecules. Through classical force field simulation, \cite{halgren1996merck} provides 
 a discrete set of rough 3D conformers (with their corresponding energies) for each molecule.
Let $\{P_i\}_{i=1}^c$ be the collection of these rough 3D conformers, and denote uniform sampling on set $\{P_i\}_{i=1}^c$ as $\textbf{Uniform}(\{P_i\}_{i=1}^c)$. Then, we can construct a discrete diffusion for the 3D structure as follows, similar to Eq. \ref{eq: closed form}:
\begin{equation}  \label{eq: discrete 3D}
P(t) = \alpha(t) \mathcal{F}_0^{-1} \cdot P(0) + \beta(t)   \textbf{Uniform}(\{\mathcal{F}_i^{-1} \cdot P_i\}_{i=1}^c),\end{equation}
where $\mathcal{F}$ is a global equivariant frame obtained by averaging the node-wise equivariant frames $\mathcal{F}_i$ defined by Eq. \ref{eq:node frame}. It can be verified that by projecting the original 3D structure with the inverse of its global frame  $\mathcal{F}^{-1}$, the above equation is $SE(3)$ invariant. 

Different from Eq. (\ref{eq: closed form}), where the randomness comes from Gaussian noise, the randomness in Eq. (\ref{eq: discrete 3D}) arises from finite uniform sampling. In conclusion, we have constructed an $SE(3)$ invariant discrete 3D diffusion based on the ground truth and rough 3D conformers.

\section{Related Works}
The primary objective of self-supervised learning (SSL) is to acquire meaningful and compressed representations from unlabeled data, often measured by their ability to estimate a specific data distribution $p_{data}$. This leads to the hypothesis that a well-trained pretrained representation can enhance the robustness and generalization of downstream tasks. However, a challenging gap exists between the ground-truth distribution $p_{data}$ (if available) and our finite set of natural data samples. To address this gap, self-supervised learning techniques such as contrastive learning and manifold learning \cite{balestriero2022contrastive} often require additional data beyond the natural samples. Contrastive learning, for example, employs data augmentation to introduce positive and negative samples, while adversarial training \cite{mathieu2016disentangling} introduces adversarial samples through data perturbation for improving the robustness of downstream classification tasks.

\paragraph{Adversarial and contrastive samples} 
In this work, we propose a diffusion framework for trajectory data augmentation based on the forward process of diffusion. A related concept that emphasizes the trajectory viewpoint is \textbf{consistency}, introduced in \cite{song2023consistency}, which requires that the points on the backward path map to the same endpoint, corresponding to the condition of the trajectories' joint distribution. Additionally, the learned reverse (generative) process of a pretrained diffusion model has been employed to generate synchronized augmentation samples \cite{trabucco2023effective}. Adversarial samples \cite{zhu2021rethinking}, compared to data augmentation, are more subtle as they involve perturbations along specific directions of the natural data that can be identified using the information provided by the score function $\nabla \log p_{data}(x)$ and $\nabla \log p_{data}(x|y)$ conditioned on labels $y$.

 Once the augmented trajectories are established, our learning objective is to fit the joint distribution of the random trajectories. From a practical standpoint, our framework lies at the intersection of two unsupervised learning paradigms: contrastive learning and generative learning. In fact, \cite{zhang2022mask} proposes an intriguing contrastive lower bound (positive sample part) for the reconstruction loss of a masked auto-encoder (AE). The authors further demonstrate that explicitly incorporating the negative sample part into the masked AE enhances the performance of the learned representation. Similarly, we observe a similar phenomenon within a more general unsupervised framework, as illustrated in Eq. \ref{absorbing}, where the masking process is realized through cold diffusion.

 
\paragraph{Molecular representation pretraining}
The labels for molecules are scarce due to the laborious and expensive cost, and self-supervised pretraining has been widely adopted to alleviate this issue.
For the \textbf{molecular topology pretraining}, existing unsupervised pretraining methods utilize either the reconstructing the masked subgraphs in molecules (AttrMask~\cite{hu2019strategies}, GPT-GNN~\cite{hu2020gpt}) or detecting the positive and negative pairs in a contrastive manner (ContextPred~\cite{hu2019strategies}, MolCLR~\cite{wang2022molecular}). Meanwhile, several works have studied the \textbf{molecular geometry pretraining}. GeoSSL~\cite{liu2023molecular} presents a comprehensive benchmark, including distance prediction, angle prediction, and an MI-based coordinate denoising framework specifically designed for conformations.
Recent progress has started to combine these two research directions, \textit{i.e.}, molecule \textbf{topology and geometry joint pretraining}. GraphMVP~\cite{liu2021pre} first proposes to maximize the MI between these two modalities in a contrastive and generative manner, and 3D InfoMax~\cite{stark20223d} is a special case of GraphMVP by only considering the contrastive part. On the other hand, Unimol~\cite{zhou2023uni} introduced a novel 3D transformer for molecular pretraining, where their pretraining tasks involve reconstructing masked topological and geometric subgraphs in combination.
The proposed \model{} in this work follows this research line.

\paragraph{Alternative training objectives of diffusion models}
In addition to the score matching training objective tailored for continuous probability distributions, an alternative training objective suitable for categorical probability distributions is the restoration loss proposed in \cite{bansal2022cold}:\cite{bansal2022cold}:
\begin{equation} \label{cold training}
\min_{\theta}\mathbb{E}_t \mathbb{E}_{x_0 \sim p_{\text{data}}} \lVert  R_{\theta}(D(x_0,t) -x_0 \rVert . 
\end{equation}
It is worth noting that compared to the score matching loss, Equation \ref{cold training} is closer to the classical auto-decoder reconstruction loss if we disregard the time variable $t$. Furthermore, \cite{li2022diffusion} has demonstrated that for discrete DDPM models, predicting $x_0$ and the score function are equivalent up to scaling constants, as the distribution of $x_{t-1}$ can be obtained in closed form through the forward process of DDPM using $x_0$ and the one-step noise.

Moreover, \cite{daras2022soft} proposed a reparameterization of the score matching loss, where the backbone neural network models the residual: $r_t = x_t - x_0$. Under the notation of Formula \ref{eq: closed form}, the Soft Score Matching (SSM) in \cite{daras2022soft} is defined as:
$$\mathcal{L}_{ssm}: = \mathbb{E}_{t \sim [0,T]} \mathbb{E}_{p(x_0)p(x_t|x_0)} [\lVert \alpha(t) (s_{\theta}(x_t,t) - r_t)\rVert^2].$$
Note that this reparameterization only works for the case when the closed form solution of the forward process has the form of Eq. \ref{eq: closed form}.

\section{Dataset Specification} \label{sec:dataset_specification}

\begin{table}[h]
\setlength{\tabcolsep}{5pt}
\fontsize{9}{9}\selectfont
\centering
\vspace{-2ex}
\caption{
\small{Some basic statistics on MD17.}
}
\label{tab:sec:MD17_statistics}
\begin{adjustbox}{max width=\textwidth}
\begin{tabular}{l r r r r r r r r}
\toprule
Pretraining & Aspirin $\downarrow$ & Benzene $\downarrow$ & Ethanol $\downarrow$ & Malonaldehyde $\downarrow$ & Naphthalene $\downarrow$ & Salicylic $\downarrow$ & Toluene $\downarrow$ & Uracil $\downarrow$ \\
\midrule
Train & 1K & 1K & 1K & 1K & 1K & 1K & 1K & 1K\\
Validation & 1K & 1K & 1K & 1K & 1K & 1K & 1K & 1K\\
Test & 209,762 & 47,863 & 553,092 & 991,237 & 324,250 & 318,231 & 440,790 & 131,770\\
\bottomrule
\end{tabular}
\end{adjustbox}
\end{table}

\section{Implementation and Hyperparameter} \label{sec:app:model_and_implementation_details}
In this section, we provide the detailed hyperparameters for implementing the experiments. 
\begin{table}[H]
\centering
\setlength{\tabcolsep}{5pt}
\fontsize{9}{9}\selectfont
\caption{
\small
Hyperparameter specifications for \model{} of property prediction.
}
\label{tab:hyperparameter}
\begin{adjustbox}{max width=\textwidth}
\begin{tabular}{l l}
\toprule
Hyperparameter & Value\\
\midrule
epochs & \{50, 100\} \\
learning rate & \{5e-4, 1e-4\} \\
$\beta$ & \{[0.1, 10]\} \\
number of steps & \{1000\}\\
$\lambda_1$ & \{1\}\\
$\lambda_2$ & \{0, 0.01, 1\}\\
\bottomrule
\end{tabular}
\end{adjustbox}
\end{table}

\begin{table}[H]
\centering
\setlength{\tabcolsep}{5pt}
\fontsize{9}{9}\selectfont
\caption{
\small
Hyperparameter specifications for \model{} of molecule generation.
}
\label{tab:hyperparametermidi}
\begin{adjustbox}{max width=\textwidth}
\begin{tabular}{l l}
\toprule
Hyperparameter & Value\\
\midrule
epochs & \{3000, 10000\} \\
learning rate & \{2e-4, 3e-4\} \\
number of layers & 12 \\
number of diffusion steps & \{500\} \\
diffusion noise schedule & cosine \\
mlp hidden dimensions & \{X: 256, E: 128, y: 128, pos: 64\} \\
$\lambda_{train}$ & \{X: 0.4, E: 2, y: 0, pos: 3, charges: 1\}\\
\bottomrule
\end{tabular}
\end{adjustbox}
\end{table}

\section{Limitations} \label{sec:app:limit}
From an algorithmic standpoint, as briefly mentioned in Section \ref{sec:reconstructive_and_contrastive_loss}, the parameterization of the score function for the reconstructive task can be obtained from the contrastive surrogate by numerically marginalizing the joint distribution $p_{\theta}(x_0,x_t)$ and subsequently applying automatic differentiation with respect to the data variable $x_0$. This methodology simplifies our two-head decoder framework described in Section \ref{sec:model_architecture} to a single head. Despite the additional computational cost associated with this approach, it holds value in terms of its theoretical significance and is therefore worth exploring in the future.

Another limitation is the treatment of the forward (noising) process of diffusion models as a form of trajectory augmentation in our current framework remains unparameterized.  As the next step, it is important to parameterize the trajectory augmentation process similar to automatic data augmentation techniques \cite{lemley2017smart, cubuk2019autoaugment}. For instance, the flexible diffusion framework proposed in \cite{du2022flexible} could be adapted to the molecule graph setting, allowing for the joint optimization of molecule (forward + backward) trajectories. By incorporating such a parameterization, we can enhance the flexibility and effectiveness of our approach in modeling realistic molecule dynamics. This extension would allow for a more fine-grained control over the trajectory augmentation process and enable the optimization of molecule dynamics in a principled manner. Exploring these possibilities represents an exciting direction for future research.

\section{More Results}
\begin{table}[h]
\setlength{\tabcolsep}{5pt}
\fontsize{9}{9}\selectfont
\centering
\vspace{-2ex}
\caption{
\small{Results on GEOM.}
}
\label{tab:sec:geom}
\begin{adjustbox}{max width=\textwidth}
\begin{tabular}{l r r r r r r r r r}
\toprule
Model & Mol stable $\uparrow$ & Atom stable $\uparrow$ & Validity $\uparrow$ & Unique $\uparrow$ & AtomTV $\downarrow$ & BondTV $\downarrow$ & ValW1 $\downarrow$ & Bond Lengths W1 $\downarrow$ & Bond Angles W1 $\downarrow$ \\
\midrule
EDM (3000 epoch) & 5.5 & 92.9 & 34.8 & 100.0 & 0.212 & 0.049 & 0.112 & 0.002 & 6.23 \\
MiDi (2D+3D, 3000 epoch) & 69.2 & 99.0 & 67.4 & 100.0 & 0.059 & 0.024 & 0.036 & 0.012 & 5.47 \\
MoleculeJAE (Pretrained on PCQM4Mv2, 569 epoch) & \textbf{84.5} & \textbf{99.6} & \textbf{79.7} & \textbf{100.0} & \textbf{0.059} & \textbf{0.021} & \textbf{0.008} & \textbf{0.003} & \textbf{2.16} \\

\bottomrule
\end{tabular}
\end{adjustbox}
\end{table}

\begin{table}[h]
\setlength{\tabcolsep}{5pt}
\fontsize{9}{9}\selectfont
\centering
\vspace{-2ex}
\caption{
\small{Results on QM9.}
}
\label{tab:sec:qm9}
\begin{adjustbox}{max width=\textwidth}
\begin{tabular}{l r r r r r r r r r}
\toprule
Model & Mol stable $\uparrow$ & Atom stable $\uparrow$ & Validity $\uparrow$ & Unique $\uparrow$ & AtomTV $\downarrow$ & BondTV $\downarrow$ & ValW1 $\downarrow$ & Bond Lengths W1 $\downarrow$ & Bond Angles W1 $\downarrow$ \\
\midrule
EDM & 90.7 & 99.2 & 91.2 & 98.5 & 0.021 & 0.002 & 0.011 & 0.001 & 0.44 \\
MiDi (2D+3D, epoch 10000) & 94.83 & 99.66 & 95.38 & \textbf{97.4} & 0.023 & 0.006 & 0.008 & 0.007 & 1.407 \\

MoleculeJAE (2D+3D, Pretrained on PCQM4Mv2, epoch 3995) & \textbf{97.42} & \textbf{99.80} & \textbf{97.65} & 97.35 & \textbf{0.006} & \textbf{0.001} & \textbf{0.002} & \textbf{0.004} & \textbf{0.561} \\

\bottomrule
\end{tabular}
\end{adjustbox}
\end{table}

\label{appendix: generation}
\paragraph{2D + 3D molecule structure generation} 
We evaluate MoleculeJAE's performance on unconditional molecule generation tasks. We generate both the graph structure and the conformer simultaneously, which is the joint distribution of the molecule's 2D and 3D structures.
This task also gives us a chance to test if our MoleculeJAE pretraining framework is compatible with the (discrete) cold diffusion framework\ref{sec:trajectory_construction}. The pretraining method still follows Figure \ref{fig:pipeline} and \ref{fig:network}, with the backbone 3D GNN of $(P(t), H(t)$ substituted by the MiDi transformer in \cite{vignac2023midi}. The discrete diffusion formulas are the same as \cite{vignac2023midi}.
\paragraph{Dataset} We still apply the PCQM4Mv2 dataset as the pretraining dataset. For the  unconditional molecule generation downstream dataset, we fine-tune our model on the QM9 dataset and the GEOM-DRUGS dataset \cite{axelrod2022geom}. GEOM comprises 430,000 drug-sized molecules with an average of 44 atoms
181 atoms. We follow \cite{vignac2023midi} to split the dataset.
\paragraph{Evaluation} We test the generation performance by calculating the probability distance of 2D and 3D structures between our generated samples with the test set. The precise definition of these metrics are provided in section 5.1 of \cite{vignac2023midi}. Besides the MiDi model in \cite{vignac2023midi}, we also use EDM \cite{hoogeboom2022equivariant} as our baseline. For our pretrained model, we finetune MoleculeJAE without the contrastive loss for 600 epoch in GEOM, and 4000 epoch in QM9.

\paragraph{Results} The experimental results are provided in Table \ref{tab:sec:geom} and \ref{tab:sec:qm9}, with our hyperparameters setting given in Table \ref{tab:hyperparametermidi}. We achieve state-of-art performance for both datasets.

\end{document}